%% file: main.tex
\documentclass{article} 
\usepackage[margin=1in]{geometry}
\usepackage{times}
\usepackage{authblk}

\input{math_commands.tex}

\usepackage{amsmath,amssymb,amsthm,mathtools}
\usepackage{bm}
\usepackage{natbib}
\usepackage{booktabs} 

\usepackage[colorlinks=true,linkcolor=blue,citecolor=blue]{hyperref}
\newcommand{\st}{\mathrm{st}}
\newcommand{\Z}{\mathbb{Z}}
\theoremstyle{plain}
\newtheorem{theorem}{Theorem}

\newtheorem{lemma}[theorem]{Lemma}

\theoremstyle{remark}

\usepackage{booktabs}
\usepackage{tabularx}
\usepackage{array}
\newcolumntype{L}[1]{>{\raggedright\arraybackslash}p{#1}}
\newcolumntype{Y}{>{\centering\arraybackslash}X}

\newcommand{\X}{\mathcal{X}}
\newcommand{\F}{\mathcal{F}}
\newcommand{\Op}[1]{\mathcal{#1}}
\newcommand{\dd}{\,\mathrm{d}}
\newcommand{\ee}{\mathrm{e}}
\newcommand{\ii}{\mathrm{i}}
\newcommand{\Ltwo}{L^2(\Omega)}

\usepackage{hyperref}
\usepackage{url}

\title{Exponential–Nemytskii Neural Operators: Learning Field Evolution Across Step Sizes}

\author[1]{Jingye Li}
\author[1]{Jinglai Li}
\affil[1]{School of Mathematics, University of Birmingham, UK}
\date{}

\begin{document}

\maketitle

\begin{abstract}
In variable-step operator learning, agreement at observed step
sizes does not by itself determine predictions at other temporal
resolutions.
We propose an operator whose step-size dependence is prescribed
by a first-order Duhamel approximation of semilinear dynamics.
It combines exact linear propagation with a local nonlinear
function of the field and its derivatives, allowing evaluation
below the observation interval without supervision densely
sampled across step sizes.
Under suitable conditions and at fixed spatial resolution,
we prove that agreement between models at two step sizes controls
their prediction differences throughout a prescribed range.
We also give conditions for local identifiability of both
components from two step sizes, supporting interpretation
of the learned structure.
A latent extension applies the same construction to learned fields.
We evaluate the base model on two synthetic problems and the
extension on two real datasets against several baselines.
\end{abstract}

\section{Introduction}
\label{sec:intro}
Many quantities of scientific and practical interest are continuous fields that evolve in space and time: the elevation of a water surface, the concentration of a diffusing species, the liquid-water content of a storm seen by radar \citep{veillette2020sevir}, or the radiative power of a spreading wildfire \citep{wooster2005}. The continuous field itself is rarely observed directly. Instead, we typically rely on a finite set of noisy samples taken on a spatial grid at discrete observation times. The task is to learn the underlying spatiotemporal evolution from these samples and to predict the field's future states.
Operator learning \citep{lu2021learning,kovachki2023neural} has emerged as a popular approach to this task. Rather than fitting a discrete mapping between fixed-size arrays, it learns the solution operator that advances the field in time as a mapping between function spaces. Because the learned operator acts on the underlying functions rather than their discrete samples, it can be evaluated on spatial grids other than those used during training, and it can be applied to complex systems even when their governing equations remain unknown. The other family of methods are equation learning \citep{brunton2019data},
which focus on recovering governing equations from the data. Our work focuses on operator
learning; equation learning is discussed further in
Section~\ref{sec:related}.

While operator learning treats space as continuous, many autoregressive
models, including FNO-2D \citep{li2021fourier}, the autoregressive version of DeepONet \citep{lu2022comprehensive} and KNO
\citep{xiong2024koopman}, advance the field at a fixed time interval
in both training and prediction.
The step size is thereby built into the learned map, which does not
directly support prediction at arbitrary temporal intervals.
Considerable effort has been devoted to developing methods for
 variable-step prediction, and we review these approaches
in Section~\ref{sec:related}.
In this work, we propose a new operator-learning method for
variable-step prediction, based on a first-order Duhamel
approximation of semilinear dynamics.
We combine this construction with multi-step-size training and
establish conditions under which prediction maps at two distinct
step sizes locally identify the linear and nonlinear components.
Our main contributions are as follows.
\begin{itemize}

\item \emph{Variable-step prediction with a stability guarantee.}
Our model's dependence on the step size is prescribed by its
mathematical construction, rather than learned entirely from data.
This structure allows us to establish a theoretical stability
guarantee for prediction across temporal resolutions, without
requiring training data densely sampled across step sizes.
We prove that, under stated conditions, close agreement with
a reference model at just two distinct training
step sizes controls prediction differences throughout a prescribed
range of step sizes, including below the observation interval (Theorem~\ref{thm:cross-step-stability}).

\item \emph{Identifiability and interpretability.}
When applied directly to the observed fields, 
our model explicitly learns a linear differential operator and
a local nonlinear function of the field and its derivatives.
These components can be inspected directly to understand how
linear propagation and nonlinear effects contribute to the prediction.
This interpretation is ambiguous when different choices of components
produce identical predictions.
We prove that, under suitable conditions, the linear and nonlinear
components of our model are locally identifiable when trained
at two distinct step sizes (Theorem~\ref{thm:identifiability}).

\end{itemize}

We test the method on two synthetic problems and on two real datasets, 
and compare it against a variety of baselines from operator
learning and equation learning.

\section{Method}
\label{sec:method}

\subsection{Setup and model derivation}
\label{sec:setup}

We observe a spatio--temporal field $u(x,t)$, $x\in\Omega\subset\R^{d_\Omega}$,
$t\in[0,T]$, only through discrete, possibly noisy samples
\begin{equation}
  \bigl\{\tilde u(x_\ell,t_j)\bigr\}_{\ell=1,\dots,N_x;\;j=1,\dots,N_t},
  \label{eq:observations}
\end{equation}
taken at $N_x$ spatial nodes $x_\ell\in\Omega$ and $N_t$ times
$t_j\in[0,T]$.
Our goal is to learn the underlying model of the field and use it to make predictions. 

We assume that $u(x,t)$ is described by a semilinear evolution equation,
\begin{equation}
  \partial_t u = A u + F(u),\qquad u(\cdot,0)=u_0,
  \label{eq:semilinear}
\end{equation}
where $A$ is an unknown linear operator generating a strongly continuous
semigroup $(\ee^{tA})_{t\ge0}$ on a Banach space $\X$, and $F:\X\to\X$ is an
unknown, locally Lipschitz nonlinear operator.  Our aim is to learn both
from Eq.~\eqref{eq:observations} alone, in the form of a neural operator that predicts $u$ at later
times.

We construct our learning model using the Duhamel representation of
Eq.~\eqref{eq:semilinear}, which does not merely split the dynamics into a
linear part and a correction, but prescribes how the two combine.
Write 
\begin{equation}
    v(t)\coloneqq\ee^{tA}u_0 \label{eq:linear}
\end{equation} for the purely linear evolution of the
initial state.  A mild solution of Eq.~\eqref{eq:semilinear} satisfies the
Duhamel formula
\begin{equation}
  u(t)\;=\;v(t)+\int_0^t \ee^{(t-s)A}F\bigl(u(s)\bigr)\dd s,
  \label{eq:duhamel}
\end{equation}
and is the fixed point of $w\mapsto v+\Op{N}[w]$, where
$\Op{N}[w](t):=\int_0^{t}\ee^{(t-s)A}F(w(s))\dd s$, constructed by Picard
iteration from $w^{(0)}\coloneqq v$.  Successive Picard iterates $w^{(k+1)}:=v+\Op{N}[w^{(k)}]$ differ by
$\Op{N}[w^{(k)}]-\Op{N}[w^{(k-1)}]$, which gains one order in $t$ at each
step since $F$ is Lipschitz, so only the first correction $\Op{N}[v]$ enters
at first order. Keeping it alone, $u\approx v+\Op{N}[v]$, and expanding
$\Op{N}[v]$ near $t=0$ (substitute $r=t-s$, then use $\ee^{rA}\approx I$ and
$v(t-r)\approx v(t)$) gives, at $t=\tau$,
\begin{equation}
  \;u(\tau)\;\approx\;v(\tau)\;+\;\tau\,F\bigl(v(\tau)\bigr).\;
  \label{eq:firstorder}
\end{equation}
Over a short horizon the solution is the {exactly propagated}
linear flow plus a single local correction evaluated at the current linear
state. Eq.~\eqref{eq:firstorder} is the
model we train to predict $u$ forward in time.

\subsection{Parametrisation and training}
\label{sec:arch}

To use Eq.~\eqref{eq:firstorder} we must learn $A$ and $F$ from data.  We
parametrise each by a structured class matching its role, rather than by a generic
neural network.

\paragraph{Spectral linear propagator.}
We assume $A$ is a constant-coefficient differential operator of order at
most $d$ on a periodic domain,
$A=\sum_{|\mathbf{n}|\le d}a_{\mathbf{n}}\partial^{\mathbf{n}}$ with
$a_{\mathbf{n}}\in\R$ and
$\partial^{\mathbf{n}}=\partial_1^{n_1}\cdots\partial_{d_\Omega}^{n_{d_\Omega}}$.
The Fourier transform $\F$ replaces $\partial_j$ by $\ii k_j$, so $A$ is the
Fourier multiplier with symbol
\begin{equation}
  g(k)\;=\;\sum_{|\mathbf{n}|\le d}c_{\mathbf{n}}\,(\ii k)^{\mathbf{n}},
  \qquad c_{\mathbf{n}}=a_{\mathbf{n}}\in\R,
  \label{eq:polysymbol}
\end{equation}
a polynomial in $\ii k$ whose coefficients are those of $A$ itself; in one
dimension $g(k)=\sum_{n\le d}c_n(\ii k)^n$.  
The linear flow Eq.~\eqref{eq:linear} becomes
$v(x,\tau)=\F^{-1}[\ee^{\tau g(k)}\widehat u_0(k)](x)$ with
$\widehat u_0=\F[u_0]$, 
and the coefficients $\{c_{\mathbf n}\}$ are the trainable parameters. 
The propagator is the \emph{exact} solution of the linear part,
so the $O(\tau^2)$ truncation in Eq.~\eqref{eq:firstorder} is attributable
to the nonlinear term alone.
In the implementation we truncate the frequency modes to $|k|\le K$, where $K$ is a hyperparameter (values in Appendix~\ref{app:hyper}).

For a bound uniform in $k$, we enforce
 $\operatorname{Re}g(k)\le\alpha$ for all $k\in\R^{d_\Omega}$,
so that $|\ee^{\tau g(k)}|\le\ee^{\alpha\tau}$: the linear flow is bounded
on $\Ltwo$ at every step size and every discretisation.
This is imposed by the following construction.  Only terms of even total
order contribute to $\operatorname{Re}g$; the odd-order terms, which carry
transport and dispersion, are purely imaginary and are left free, as is the
constant term $c_{\mathbf{0}}=:\alpha$.  Among the even-order terms, those in which every $k_j$ appears to an even
power ($k_1^2$, $k_1^2k_2^2$, and so on) are nonnegative everywhere and
receive a nonpositive coefficient, so they can only push $\operatorname{Re}g$
down; the rest, such as $k_1k_2$, change sign with direction and are dropped.

\paragraph{Nemytskii operator.}
For the nonlinear part of Eq.~\eqref{eq:semilinear} we do not consider
nonlocal nonlinearities, and assume instead that $F$ is local of some order
$\rho\in\mathbb N_0$: the value of $F(u)$ at $x$ depends on $u$ only through
$u(x)$ and its spatial derivatives at $x$ up to order $\rho$ (the
$\rho$-jet of $u$ at $x$). Write
\begin{equation}
  \Phi[w](x)\;:=\;\bigl(w(x),\,\nabla w(x),\,\dots,\,\nabla^{\rho}w(x),\,\eta(x)\bigr)\;\in\;\R^{m}
  \label{eq:featuremap}
\end{equation}
for the \emph{feature map}, which collects the field, its spatial derivatives
up to order $\rho$, and, optionally, externally given fields $\eta(x)$; its
number of entries is $m$.  We then assume
\begin{equation}
  F(u)(x)\;=\;f\bigl(\Phi[u](x)\bigr),\qquad f:\R^{m}\to\R,
  \label{eq:nemytskii}
\end{equation}
so that $F$ is a generalised Nemytskii (superposition) operator
\citep{li2025nno} induced by the finite-dimensional map $f$.
Accordingly, the goal becomes to learn the finite-dimensional 
function $f:\R^{m}\to\R$ and we represent it by a multilayer perceptron (MLP) with smooth
activations.  
The derivatives in $\Phi$ are evaluated spectrally,
$\nabla^{j}v=\F^{-1}[(\ii k)^{j}\widehat v]$.

\paragraph{Exponential-Nemytskii neural operator.}
Combining the two components using Eq.~\eqref{eq:firstorder},
\begin{equation}
  \Op{S}_\tau[u_0](x)
  \;=\;
  \underbrace{\F^{-1}\bigl[\ee^{\tau g}\F u_0\bigr](x)}_{\textstyle v(x,\tau)}
  \;+\;\tau\,f\bigl(\Phi[v(\cdot,\tau)](x)\bigr),
  \label{eq:onestep}
\end{equation}
with trainable objects $g$ and $f$.
We call the resulting model the \emph{Exponential--Nemytskii Neural
Operator} (ENNO), reflecting its combination of exponential linear
propagation and a learned Nemytskii map acting on local
field features.
Predictions over horizons longer than a single first-order step are
obtained by applying Eq.~\eqref{eq:onestep} autoregressively:
starting from $u_0$,
\begin{equation}
  u_{j+1}\;=\;\Op{S}_\tau[u_j],\qquad j=0,\dots,n-1,
  \qquad\text{so that}\qquad
  u_n\;=\;\Op{S}_\tau^{\,n}[u_0]
  \label{eq:rollout}
\end{equation}
is the prediction at time $t=n\tau$.

\paragraph{Training.}
In training we use only the one-step model Eq.~\eqref{eq:onestep}; the
multi-step composition Eq.~\eqref{eq:rollout} is used only for
prediction.
A key feature of our approach is that the model is trained with several step
sizes $\tau$, which is required for the identifiability of $(g,f)$
(Section~\ref{sec:identifiability}).
Concretely, fix step sizes $\tau_1<\dots<\tau_S$, $S\ge2$, chosen among the
differences of observation times, and for each $\tau_s$ let $\Op{D}_s$ be
the set of observation times $t_j$ for which $t_j+\tau_s$ is also an
observation time.  For each $t_j\in\Op{D}_s$ we advance $\tilde u(\cdot,t_j)$
by one step of size $\tau_s$ and compare with the observation at
$t_j+\tau_s$; averaging the squared error over the grid and over $\Op{D}_s$
gives
\begin{equation}
  \mathcal{L}_s
  \;=\;
  \frac{1}{|\Op{D}_s|}\sum_{t_j\in\Op{D}_s}
  \frac{1}{N_x}\sum_{\ell=1}^{N_x}
  \bigl|\,\Op{S}_{\tau_s}\bigl[\tilde u(\cdot,t_j)\bigr](x_\ell)
  -\tilde u(x_\ell,t_j+\tau_s)\,\bigr|^2 ,
  \label{eq:loss-s}
\end{equation}
and the training objective is their sum, $\mathcal{L}=\sum_{s=1}^{S}\mathcal{L}_s$.
Since $\tau$ enters $\Op{S}_\tau$ explicitly, the same $(g,f)$ is used for
every $s$.  With a single step size the constant term of $g$ and the linear
part of $f$ cannot be separated (Lemma~\ref{lem:single-step}); a second
step size removes this ambiguity (Theorem~\ref{thm:identifiability}(ii)).

\paragraph{Latent fields.}
An extension of the method handles the case where $u(x,t)$ does not follow
the semilinear model Eq.~\eqref{eq:semilinear}: we map $u(x,t)$ into a
latent field, or several latent fields, using an encoding--decoding
structure.
Namely, we use an encoder $\Op{E}$ to map the observed history into the
latent fields
$w=(w^{1},\dots,w^{M})$, advance all of them with the same $g$ and $f$, and
use a decoder $\Op{R}$ to map the latent state back:
\begin{equation}
  w_0=\Op{E}[u_{-T+1},\dots,u_0],\qquad
  w_{j+1}=\bigl(\Op{S}_\tau[w_j^{1}],\dots,\Op{S}_\tau[w_j^{M}]\bigr),\qquad
  u_j=\Op{R}[w_j].
  \label{eq:latentrollout}
\end{equation}
Here we restrict both $\Op{E}$ and $\Op{R}$ to be Nemytskii operators,
generated by pointwise maps $\R^{T}\to\R^{M}$ and $\R^{M}\to\R$, so every
learnable object in the model is generated by a map between
finite-dimensional spaces.  The encoder and decoder are trained jointly
with $g$ and $f$.
Taking $T>1$ acts as a delay embedding: when $u$ is a partial observation of
a higher-dimensional state, a single frame need not determine the subsequent
evolution.  Only the initial latent state uses the history; the rollout
remains Markovian in $w$.
Note that $g$ and $f$ are shared across the latent fields, so the parameter
count does not grow with $M$: the $M$ fields evolve independently under the
same dynamics but from different initial channels produced by $\Op{E}$, and
are recombined nonlinearly by $\Op{R}$. 
The price is interpretability: in the base model $g$ and $f$ describe the
mechanisms governing $u$ itself, whereas here they describe the latent
fields and no longer correspond directly to the equation for $u$.

\subsection{Theory}

\subsubsection{Stability across step sizes}
\label{sec:stability}

We work at fixed spatial resolution $K$ on a periodic box $\Omega$.
Let $\Op{V}_K$ be the real Fourier space with integer mode indices
$|k|\le K$, let $P_K$ be its orthogonal projection, and write
$\Op{U}_R:=\{u\in\Op{V}_K:\|u\|_{L^2}\le R\}$.
All operator norms below are induced by the $L^2$ norm.
For $\vartheta=(g,f)$, let $A$ have symbol $g$ as in
Eq.~\eqref{eq:polysymbol}, and set
$F_K(v)=P_K[f(\Phi_\st[v])]$ and
$\Op{S}^{K,\vartheta}_t u
=\ee^{tA}u+tF_K(\ee^{tA}u)$.
Here $f$ is a spatially homogeneous, locally Lipschitz scalar
function, and $\Phi_\st[v]$ collects the field and its distinct
partial derivatives through order $\rho\ge0$.
For a second pair $\widetilde\vartheta=(\widetilde g,\widetilde f)$
from the same class, define $\widetilde A$, $\widetilde F_K$,
and $\Op{S}^{K,\widetilde\vartheta}_t$ analogously.
Let $\mathcal M_K$ denote the real-preserving Fourier multipliers
on $\Op{V}_K$, and define the nonlinear contrast
$\mathcal N_F(v):=F_K(v)-2F_K(v/2)+F_K(0)$.
We have the following theorem where the proof is given 
in Appendix \ref{app:stability}.

\begin{theorem}[Cross-step stability]
\label{thm:cross-step-stability}
Fix $R,\Delta,M>0$, and suppose
$\|A\|_{\mathrm{op}},\|\widetilde A\|_{\mathrm{op}}\le M$.
Assume that there exist $0<r_0<R$ and $\sigma,\zeta>0$ such that
\begin{equation}
  \sup_{v\in\Op{U}_{r_0}}
  \|\mathcal N_F(\ee^H v)-\mathcal N_F(v)\|_{L^2}
  \ge \sigma\|H\|_{\mathrm{op}},
  \qquad
  H\in\mathcal M_K,\quad \|H\|_{\mathrm{op}}\le\zeta.
  \label{eq:nonlinear-separation}
\end{equation}
Let $L_\Delta$ be a Lipschitz constant for $F_K$ on
$\Op{U}_{\exp(2M\Delta)R}$.
There exist $0<\tau_\ast\le\Delta$, $0<r_\Delta<R$, and
$C_\Delta>0$, depending only on
$\Omega,K,M,R,\Delta,r_0,\sigma,\zeta,L_\Delta$,
such that for any $0<\tau_0<\tau_1\le\tau_\ast$, setting
\[
  \varepsilon_{\rm obs}:=
  \max_{j=0,1}\ \sup_{u\in\Op{U}_R}
  \frac{
    \|\Op{S}^{K,\vartheta}_{\tau_j}u
      -\Op{S}^{K,\widetilde\vartheta}_{\tau_j}u\|_{L^2}
  }{\tau_j},
\]
we have, for every $0<\delta\le\Delta$,
\begin{equation}
  \sup_{u\in\Op{U}_{r_\Delta}}
  \frac{
    \|\Op{S}^{K,\vartheta}_\delta u
      -\Op{S}^{K,\widetilde\vartheta}_\delta u\|_{L^2}
  }{\delta}
  \le
  C_\Delta\varepsilon_{\rm obs}
  \left(
    1+\frac{2\min_{\tau\in[\tau_0,\tau_1]}|\delta-\tau|}
    {\tau_1-\tau_0}
  \right).
  \label{eq:cross-step-stability}
\end{equation}
The linear dependence on this ratio is sharp in general.
\end{theorem}

\paragraph{Interpretation.}
The theorem can be understood as follows.
If a model's predictions are close to those of a reference model
at two training step sizes, their prediction difference per unit
time also has an explicit upper bound at other step sizes.
At intermediate step sizes, the same theoretical bound applies
as at the training steps.
Outside this interval, the bound grows linearly with the
extrapolation distance relative to the separation between the
training steps.
This provides a stability guarantee for using the same model
at different temporal resolutions. 

\subsubsection{Identifiability}
\label{sec:identifiability}
We study structural identifiability of  $(g,f)$ from the one-step
prediction maps.
Write $\Phi[\phi]=(\Phi_\st[\phi],\eta)$, where $\Phi_\st$ lists each
distinct partial derivative through order $\rho$ once, including the
field value. Accordingly, write $z=(z_\st,z_\eta)$, with $z_0$ the
field value. Let $\eta:\Omega\to\R^{d_\eta}$ be continuous and
$E:=\eta(\Omega)$.
Proofs of the two results below are in Appendix~\ref{app:identifiability}.

\begin{lemma}[Single-step degeneracy]
\label{lem:single-step}
Fix $\tau_0>0$. For $\gamma\in\R$, set
$\varrho_\gamma:=\ee^{\gamma\tau_0}$ and
$\chi_\gamma:=(1-\varrho_\gamma)/(\varrho_\gamma\tau_0)$,
and define
\begin{equation}
  g_\gamma:=g+\gamma,\qquad
  f_\gamma(z_\st,z_\eta):=
  f(z_\st/\varrho_\gamma,z_\eta)+\chi_\gamma z_0.
  \label{eq:fgamma}
\end{equation}
Then
\begin{equation}
  \Op{S}_{\tau_0}[\phi;g_\gamma,f_\gamma]
  =\Op{S}_{\tau_0}[\phi;g,f]
  \label{eq:degeneracy}
\end{equation}
whenever the maps are defined.
\end{lemma}

\begin{theorem}[Identifiability]
\label{thm:identifiability}
Let $(g,f)$ and $(g',f')$ have symbols in the real-coefficient
polynomial class of Eq.~\eqref{eq:polysymbol} and nonlinearities that
are $C^1$ in $z_\st$ and continuous in $z_\eta$.
Fix $r>0$ and $K\ge K_0(d,\rho,d_\Omega)$, with the sufficient
threshold $K_0$ specified in Appendix~\ref{app:identifiability}.
At every step size $\tau$ considered, assume
$\tau\,|\operatorname{Im}\widetilde g(k)|<\pi/4$ for
$\widetilde g\in\{g,g'\}$ and all $k\in\Z^{d_\Omega}$ with $|k|\le K$.
\begin{enumerate}\itemsep3pt
\item[(i)] \emph{One step size.}
Let $\tau_0>0$. Suppose
$\Op{S}_{\tau_0}[\phi;g,f]=\Op{S}_{\tau_0}[\phi;g',f']$ on $\Omega$
for every $\phi\in\Op{V}_K$ with $\|\phi\|_{L^2}<r$, and
$1+\tau_0\,\partial_{z_0}f(0,\eta(x_0))\ne0$ for some $x_0\in\Omega$.
Then, for some $\gamma\in\R$ and $r_\ast>0$,
$g'=g+\gamma$ and $f'=f_\gamma$ on $B_{r_\ast}(0)\times E$ with
$B_{r_\ast}(0)\subset\R^{m_\st}$, where $f_\gamma$ is defined by
Eq.~\eqref{eq:fgamma}.
In particular, $c'_{\mathbf n}=c_{\mathbf n}$ for $1\le|\mathbf n|\le d$.

\item[(ii)] \emph{Two step sizes.}
Under the hypotheses of {\rm(i)}, suppose also
$\Op{S}_{\tau_1}[\phi;g,f]=\Op{S}_{\tau_1}[\phi;g',f']$ on $\Omega$
for the same $\phi$, where $\tau_1>0$, $\tau_1\ne\tau_0$.
If, for some $z_\eta^\ast\in E$, on no neighbourhood of the origin
is $f(\cdot,z_\eta^\ast)$ of the form $a+bz_0$ for constants
$a,b\in\R$, then $\gamma=0$.
Thus $g'=g$, and $f'=f$ on $B_{r_\ast}(0)\times E$.
\end{enumerate}
\end{theorem}

\paragraph{Interpretation.}
Lemma \ref{lem:single-step} shows that the linear and nonlinear components cannot
be uniquely separated at a single step size, and longer rollouts
at that same step size do not resolve this ambiguity.
Theorem \ref{thm:identifiability} states that 
a second distinct step size removes this ambiguity under
the stated conditions.
Thus, using two distinct step sizes resolves the ambiguity between
the linear and nonlinear components, establishing structural
identifiability of the model.

\subsection{Related work}
\label{sec:related}
We review existing methods that support variable-step prediction.
Equation-learning methods can recover continuous-time governing
equations from data
\citep{brunton2016discovering,rudy2017data,long2018pde}.
These equations permit flexible time stepping, although prediction
still relies on numerical integration of the learned dynamics.
Here we focus on operator-learning methods, which learn evolution
maps directly.
One such approach treats time as a query coordinate, as in DeepONet
with a space--time trunk \citep{lu2021learning} and nonlinear
coordinate decoders \citep{seidman2022nomad}.
These models directly learn spatiotemporal neural representations and may require substantial paired training data \citep{wang2021learning}.
The second approach, exemplified by Poseidon \citep{herde2024poseidon}
and OSG-FNO \citep{chen2025due}, takes the step size $\tau$ as an input.
Such models can in principle be evaluated at any step size, but
their dependence on $\tau$ is entirely learned through training.
Outside the range of observed step sizes---in particular below
the finest step the data provide---fitting the available data
alone cannot guarantee reliable prediction.
A third approach learns a vector field in function space and
predicts by numerical integration: TI-DeepONet
\citep{nayak2026ti} fits it through a numerical integrator,
while CFO \citep{hou2026cfo} fits it to estimated time derivatives.
The learned dynamics may consequently inherit errors from the
numerical integration or derivative estimation used during training.

In our numerical experiments we compare against one representative of each of
the three families above: NOMAD \citep{seidman2022nomad} for temporal-spatial representation, OSG-FNO \citep{chen2025due} for step-size learning, and
TI-DeepONet \citep{nayak2026ti} for learned vector fields.
We further compare against E-WSINDy
\citep{fasel2022esindy} from the equation-learning
literature, KNO \citep{xiong2024koopman} as a fixed-step 
method, and persistence.

\section{Numerical experiments}
We present four numerical experiments: two synthetic examples and two applications to real-world data.
We compare our method with 5 benchmarks discussed in Section \ref{sec:related}. 
We use the recommended or default settings from the original
papers and implementations whenever available and applicable.
Otherwise, we select suitable hyperparameters based on
validation performance.
The architecture, spectral cutoff, symbol degree and data settings of ENNO in each experiment are listed in Appendix~\ref{app:hyper}.
All neural models are trained for at most 3,000 epochs.
All baselines are configured to a parameter count comparable to ours, except OSG-FNO, which we could not reduce below three times ours without seriously degrading its training.

\subsection{The KdV equation}

The first synthetic example is the Korteweg--de Vries equation
$u_t+6uu_x+u_{xxx}=0$ on the periodic domain $[0,10]$.
Its symbol $g(k)=\ii k^{3}$ lies in the polynomial class of
Eq.~\eqref{eq:polysymbol} once $d\ge3$, with $c_3=-1$ and all other
coefficients zero, so the learned symbol can be compared with the truth
coefficient by coefficient.
Training uses 2030 trajectories on an $N_x=128$ grid, corrupted by Gaussian
noise at $5\%$ of each trajectory's own standard deviation; 50 and 200
further trajectories with unseen initial conditions serve for validation and
test.  The  model is
trained with two step sizes, $\tau_1=10^{-3}$ and
$\tau_2=2\times 10^{-3}$.  KNO needs ten frames of context, so all methods are scored on the same
targets, $u(t_1),u(t_5),u(t_{10})$; the single-frame methods are conditioned
on $u(t_0)$ and KNO on the ten frames up to and including $u(t_0)$.

\begin{table}[!ht]
  \centering
  \small
  \caption{KdV: relative $L^2$ error (\%) at three target frames, averaged
  over the 200 test trajectories, for symbol degrees $d=1,\dots,4$.}
  \label{tab:kdv_degree}
  \begin{tabular}{@{}lrrr@{}}
    \toprule
    & $t_{1}$ & $t_{5}$ & $t_{10}$ \\
    \midrule
    $d=1$ & 2.895 & 3.676 & 4.200 \\
    $d=2$ & 2.179 & 2.695 & 2.989 \\
    $d=3$ & \textbf{0.313} & \textbf{0.721} & \textbf{1.115} \\
    $d=4$ & 0.319 & 0.743 & 1.151 \\
    \bottomrule
  \end{tabular}
\end{table}

The degree $d$ decides whether the true model can be recovered, so we first
ablate it in Table~\ref{tab:kdv_degree}, with everything else held fixed.
The error falls by a factor of $2.7$ from $d=2$ to $d=3$ and does not improve
clearly at $d=4$.  We use $d=3$ in what follows.  We also
ablate the latent-field structure, which brings no improvement here, as
expected since the true model lies in our class at this degree
(Appendix~\ref{app:kdv}).
To test the identifiability of the two components, we show the recovered
coefficients in Table~\ref{tab:kdv_coeff}, where all of them agree well with
the truth.  Table~\ref{tab:kdv} compares against the baselines, reporting the mean over the 200 test trajectories with its standard error of the mean (SE); ENNO is lowest at all four target times.  The $t_{1/2}$ column evaluates each model
at half the smallest training step, which none of them has seen: KNO cannot
be evaluated there at all, and OSG-FNO and TI-DeepONet are worse at
$t_{1/2}$ than at $t_1$ despite the shorter horizon.
\begin{table}[!ht]
  \centering
  \small
  \caption{KdV: recovered symbol coefficients at $d=3$ against the truth.}
  \label{tab:kdv_coeff}
  \begin{tabular}{@{}lrrrr@{}}
    \toprule
    & $c_0$ & $c_1$ & $c_2$ & $c_3$ \\
    \midrule
    truth     & $0$ & $0$ & $0$ & $-1$ \\
    recovered & $-0.061$ & $-0.081$ & $4.9\times10^{-5}$ & $-0.979$ \\
    \bottomrule
  \end{tabular}
\end{table}

\begin{table}[!ht]
  \centering
  \small
  \setlength{\tabcolsep}{5pt}
  \caption{KdV: relative $L^2$ error (\%) at four target times, reported as
  mean $\pm$ SE over the 200 test trajectories. $t_{1/2}$ is half
  the smallest training step and is not an observation time; methods that
  cannot be evaluated below the observation interval are marked ``---''.}
  \label{tab:kdv}
  \begin{tabular}{@{}lrrrr@{}}
    \toprule
    Method & $t_{1/2}$ & $t_1$ & $t_5$ & $t_{10}$ \\
    \midrule
    ENNO ($d=3$) & \textbf{0.197} $\pm$ 0.008 & \textbf{0.313} $\pm$ 0.015 & \textbf{0.721} $\pm$ 0.032 & \textbf{1.115} $\pm$ 0.049 \\
    TI-DeepONet  & 2.374 $\pm$ 0.066 & 1.369 $\pm$ 0.035 & 2.556 $\pm$ 0.078 & 3.329 $\pm$ 0.105 \\
    E-WSINDy     & 0.830 $\pm$ 0.061 & 1.459 $\pm$ 0.079 & 2.436 $\pm$ 0.124 & 2.793 $\pm$ 0.143 \\
    OSG-FNO      & 3.264 $\pm$ 0.083 & 2.045 $\pm$ 0.089 & 2.008 $\pm$ 0.088 & 2.112 $\pm$ 0.080 \\
    KNO          & --- & 2.049 $\pm$ 0.096 & 1.824 $\pm$ 0.083 & 1.912 $\pm$ 0.087 \\
    NOMAD        & 4.800 $\pm$ 0.147 & 4.815 $\pm$ 0.152 & 4.804 $\pm$ 0.152 & 4.832 $\pm$ 0.149 \\
    \bottomrule
  \end{tabular}
\end{table}

\subsection{The advection-diffusion equation}

The second synthetic example is the advection-diffusion equation
$u_t=\nu(u_{xx}+u_{yy})-u(u_x+u_y)$ with $\nu=0.05$ on the periodic domain
$[0,1]^2$.  Its symbol $g(\mathbf k)=-\nu|\mathbf k|^{2}$ lies in the
polynomial class of Eq.~\eqref{eq:polysymbol} once $d\ge2$, with
$c_{(2,0)}=c_{(0,2)}=\nu$ and all other coefficients zero, so the learned
symbol can again be compared with the truth coefficient by coefficient.
Training uses 100 trajectories on a $128\times128$ grid, corrupted by
Gaussian noise at $5\%$ of each trajectory's own standard deviation; 10 and
25 further trajectories serve for validation and test.  The encoder and
decoder are identity maps, and the model is trained with two step sizes,
$\tau_1=10^{-3}$ and $\tau_2=2\times 10^{-3}$.  As on KdV, all methods use the last input
frame only, except KNO which uses  the preceding ten.

The ablation in Table~\ref{tab:adv_degree} selects $d=2$, the true order,
with no clear improvement at $d=3$.
We first estimate the symbol coefficients: the diffusion coefficient is
recovered as $0.0488$ against the true $0.05$, and the remaining
coefficients, whose true values are zero, stay below $0.07$ in magnitude
(Table~\ref{tab:adv_coeff}).
Table~\ref{tab:adv} compares against the baselines at comparable parameter
counts.  ENNO is lowest at all four target times, with E-WSINDy the closest
competitor, unlike on KdV, where its candidate library must cover a
third-order dispersive term.  At $t_{1/2}$ it remains lowest, while KNO
cannot be evaluated there at all.  
Neither branch--trunk method tracks the dynamics here: the error of NOMAD is already at the scale of the solution at $t_1$ and barely changes over the horizon, while that of TI-DeepONet grows from about $15\%$ at $t_1$ to $51\%$ at $t_{10}$. Since TI-DeepONet is competitive on KdV (second-best at $t_1$), the failure here is not one of training budget.

\begin{table}[!ht]
  \centering
  \small
  \caption{Advection--diffusion: relative $L^2$ error (\%) at three target
  frames, averaged over the 25 test trajectories, for symbol degrees
  $d=1,2,3$.}
  \label{tab:adv_degree}
  \begin{tabular}{@{}lrrr@{}}
    \toprule
    & $t_{1}$ & $t_{5}$ & $t_{10}$ \\
    \midrule
    $d=1$ & 2.439 & 4.011 & 4.647 \\
    $d=2$ & \textbf{0.021} & \textbf{0.074} & \textbf{0.129} \\
    $d=3$ & 0.021 & {0.074} & 0.128 \\
    \bottomrule
  \end{tabular}
\end{table}

\begin{table}[htbp]
\centering
\small
\caption{Advection--diffusion: recovered symbol coefficients at $d=2$ against
the truth. The two first-order coefficients are listed as $x/y$.}
\label{tab:adv_coeff}
\begin{tabular}{@{}lrrr@{}}
\toprule
& $c_{(0,0)}$ & $c_{(1,0)}/c_{(0,1)}$ & $c_{(2,0)}=c_{(0,2)}$ \\
\midrule
truth     & $0$     & $0/0$         & $0.05$ \\
recovered & $0.029$ & $0.054/0.061$ & $0.0488$ \\
\bottomrule
\end{tabular}
\end{table}

\begin{table}[!ht]
  \centering
  \small
  \setlength{\tabcolsep}{5pt}
  \caption{Advection--diffusion: relative $L^2$ error (\%) at four target
  times, reported as mean $\pm$ SE over the 25 test trajectories.
  $t_{1/2}$ is half the smallest training step and is not an observation
  time; methods that cannot be evaluated below the observation interval are
  marked ``---''.}
  \label{tab:adv}
  \begin{tabular}{@{}lrrrr@{}}
    \toprule
    Method & $t_{1/2}$ & $t_{1}$ & $t_{5}$ & $t_{10}$ \\
    \midrule
    ENNO ($d=2$) & \textbf{0.011} $\pm$ 0.001 & \textbf{0.021} $\pm$ 0.002 & \textbf{0.074} $\pm$ 0.008 & \textbf{0.129} $\pm$ 0.013 \\
    E-WSINDy     & 0.030 $\pm$ 0.003 & 0.030 $\pm$ 0.003 & 0.145 $\pm$ 0.016 & 0.283 $\pm$ 0.030 \\
    KNO          & --- & 0.269 $\pm$ 0.037 & 0.376 $\pm$ 0.049 & 0.562 $\pm$ 0.072 \\
    OSG-FNO      & 0.262 $\pm$ 0.037 & 0.524 $\pm$ 0.060 & 2.617 $\pm$ 0.303 & 5.223 $\pm$ 0.607 \\
    TI-DeepONet  & 14.597 $\pm$ 2.122 & 14.551 $\pm$ 2.068 & 29.970 $\pm$ 3.330 & 50.831 $\pm$ 5.843 \\
    NOMAD        & 79.330 $\pm$ 5.491 & 60.124 $\pm$ 4.971 & 59.634 $\pm$ 4.909 & 58.969 $\pm$ 4.898 \\
    \bottomrule
  \end{tabular}
\end{table}

\subsection{Radar nowcasting}
\label{sec:sevir}
Radar nowcasting aims to predict the precipitation field 0--2 hours ahead from recent radar observations, filling the gap that numerical weather prediction leaves at these lead times.
We take the vertically integrated liquid (VIL) product of SEVIR \citep{veillette2020sevir} as the
observed field $u(x,t)$, sampled on a uniform $128\times128$ grid over a
square domain $\Omega\subset\R^{2}$ of $384\times384\ \mathrm{km}^{2}$,
with consecutive frames five minutes apart.
Given the ten most recent frames, the task is to predict the next ten frames
at five-minute intervals, covering lead times of 5--50 minutes.
We use 11,159 training events, 200 validation events and 500 test
events; the test events come from separate files covering
November--December 2019.
Our model encodes the ten input frames into 20 latent fields and is trained
with two step sizes, $\tau_1=5$ and $\tau_2=10$ minutes.
All methods predict the same final ten frames of each test event.

\begin{table}[!ht]
  \centering
  \small
  \setlength{\tabcolsep}{6pt}
  \caption{SEVIR test MSE ($\times 10^{-3}$) at three forecast lead times,
  reported as mean $\pm$ SE over the 500 test events. Bold marks
  the best result. ``Failed'' denotes nonfinite predictions during
  rollout.}
  \label{tab:real_sevir}
  \begin{tabular}{@{}lrrr@{}}
    \toprule
    Method & 5 minutes & 25 minutes & 50 minutes \\
    \midrule
    ENNO          & \textbf{1.4} $\pm$ 0.05 & \textbf{5.4} $\pm$ 0.15 & \textbf{9.2} $\pm$ 0.25 \\
    KNO           & 1.8 $\pm$ 0.05 & 6.7 $\pm$ 0.17 & 10.3 $\pm$ 0.28 \\
    OSG-FNO       & 1.8 $\pm$ 0.04 & 8.6 $\pm$ 0.15 & 13.7 $\pm$ 0.26 \\
    NOMAD         & 5.6 $\pm$ 0.14 & 8.9 $\pm$ 0.24 & 13.4 $\pm$ 0.40 \\
    TI-DeepONet   & 6.4 $\pm$ 0.16 & 11.4 $\pm$ 0.29 & 19.5 $\pm$ 0.50 \\
    E-WSINDy      & 12.5 $\pm$ 0.30 & Failed & Failed \\
    \midrule
    Persistence   & 2.6 $\pm$ 0.09 & 15.3 $\pm$ 0.42 & 24.3 $\pm$ 0.66 \\
    \bottomrule
  \end{tabular}
\end{table}

\begin{figure}[t]
  \centering
  \includegraphics[width=0.65\linewidth]{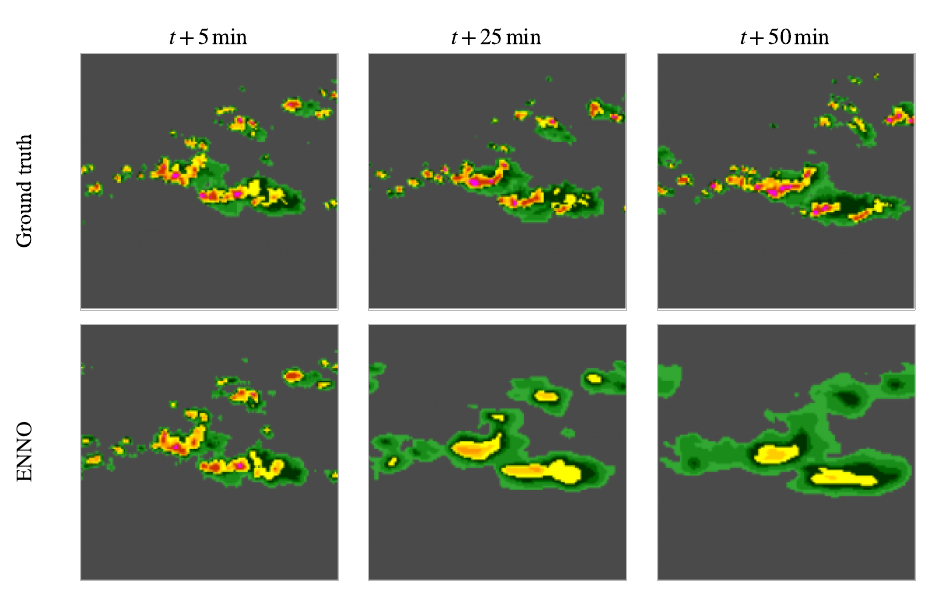}
\caption{Qualitative SEVIR forecasts for one test event at lead times of
5, 25 and 50 minutes, matching Table~\ref{tab:real_sevir}. Top row: ground
truth; bottom row: ENNO's autoregressive predictions.}
  \label{fig:sevir_qualitative}
\end{figure}

Our model achieves the lowest MSE at all three lead times in
Table~\ref{tab:real_sevir}.
The comparison with KNO is particularly informative because both
models receive the same ten-frame history.
Relative to KNO, our model reduces MSE by about 22\%, 19\% and 11\% at 5, 25 and 50 minutes, respectively.
The improvement is therefore present in the first prediction and
persists at all three lead times, demonstrating a sustained
advantage over the tested forecast horizon.
Figure~\ref{fig:sevir_qualitative} illustrates the rollout on a test event.
ENNO tracks the motion and coarse structure of the storm cells throughout,
while the high-intensity cores are progressively smoothed.

\subsection{Wildfire prediction}
\label{sec:wildfire}

Wildfire management relies on knowing where a fire will burn next and how
intensely.
The goal of this example is to predict the next day's fire radiative power (FRP) field from the FRP
observed on the preceding days together with environmental drivers---wind,
terrain, vegetation and moisture---recorded on the most recent day.
We use WildfireSpreadTS~\citep{Gerard2023WildfireSpreadTSAD}, which records 607
wildfire events across the western United States between 2018 and 2021 as
daily multi-channel images at $375$\,m resolution, one channel holding the observed FRP and the rest holding environmental drivers for the same day.
We take $T_{\rm in}\in\{1,5\}$ daily FRP fields as the state, transformed by
$\log(1+\mathrm{FRP})$ since FRP spans about four orders of magnitude, and 25
environmental channels from the most recent input day.
The latter enter the model as the externally given fields $\eta$ of
Eq.~\eqref{eq:featuremap}, so that $f$ can modulate the predicted intensity
pointwise according to local wind, slope and fuel conditions.
Our model is trained with two step sizes, $\tau_1=1$ and $\tau_2=2$ days;
evaluation is always at one day ahead.
All neural predictors are trained with the same weighted squared loss on
log-FRP, giving weight 50 to pixels with positive FRP and weight 1 to the
remaining valid pixels: fire pixels make up under $0.15\%$ of the valid
pixels, and without this weighting training can favour near-zero predictions.
Pixels introduced by padding events to a common grid size are excluded from
both training and evaluation.

We use four folds over 2018--2021, one per test year. Each fold trains on two
years, selects a checkpoint on a third, and evaluates on the remaining year.
Model selection and evaluation use the area under the precision--recall curve
(AUC-PR), with predicted log-FRP serving as a continuous score for detecting
pixels with positive FRP.
Table~\ref{tab:real_wsts} shows that our model achieves the highest
mean AUC-PR at both history lengths.
With one input day, it leads in every held-out year, attaining a
mean of 0.3210 compared with 0.2895 for KNO.
With five input days, our model and OSG-FNO achieve similar means (0.3122 versus 0.3098), each leading in two of the four held-out years; OSG-FNO uses more than ten
times as many trainable parameters ($8.4$M versus $0.75$M).
Longer histories do not improve our model's mean AUC-PR here,
and persistence exceeds three of the trained baselines with
one input day.

\begin{table}[!ht]
  \centering
  \small
  \setlength{\tabcolsep}{6pt}
  \caption{WildfireSpreadTS test AUC-PR, reported as mean
  $\pm$ sample standard deviation across four held-out years.
  Bold marks the highest mean.
  ``Unsupported'' indicates that the evaluated implementation
  does not support five-frame inputs.}
  \label{tab:real_wsts}
  \begin{tabular}{@{}lcc@{}}
    \toprule
    & \multicolumn{2}{c}{AUC-PR} \\
    \cmidrule(lr){2-3}
    Method & $T_{\rm in}=1$ & $T_{\rm in}=5$ \\
    \midrule
    ENNO
      & $\mathbf{0.3210}\pm0.0848$
      & $\mathbf{0.3122}\pm0.0835$ \\
    KNO
      & $0.2895\pm0.0867$
      & $0.2889\pm0.0744$ \\
    OSG-FNO
      & $0.2664\pm0.0811$
      & $0.3098\pm0.0887$ \\
    TI-DeepONet
      & $0.1435\pm0.0373$
      & Unsupported \\
    NOMAD
      & $0.1290\pm0.0394$
      & Unsupported \\
    E-WSINDy
      & $0.0337\pm0.0635$
      & $0.2533\pm0.0954$ \\
    \midrule
    Persistence
      & $0.2473\pm0.0984$
      & $0.2530\pm0.0972$ \\
    \bottomrule
  \end{tabular}
\end{table}

\section{Conclusions}
\label{sec:conclusions}
We presented ENNO, an operator-learning method that combines exact
linear propagation with a local nonlinear correction and prescribes
its step-size dependence through a first-order Duhamel approximation.
Under suitable conditions, prediction maps at two distinct step sizes
locally identify these components, and agreement between models at
those steps controls prediction differences across a prescribed range.
Experiments on two synthetic PDEs and two real-world datasets
demonstrate strong predictive performance, with a latent extension
handling the real-data applications.
The synthetic examples also show accurate recovery of the principal
differential coefficients.
The current guarantees concern the base model at fixed spatial
resolution. Extending the analysis to the latent construction
and obtaining stability bounds uniform under spatial refinement
are left to future work.

\newpage

\subsection*{AI use statement}

In this work, we used generative AI tools for tasks that require disclosure as
follows. The theoretical results of this paper---the model construction, the
statements of Theorem~\ref{thm:cross-step-stability},
Lemma~\ref{lem:single-step} and Theorem~\ref{thm:identifiability}, and the
proof strategies---were conceived and developed in full by the authors, and
complete proofs were written before any AI tool was involved. We then used
generative AI tools to check these proofs. The tools identified a number of
errors and proposed corrections, some of which we adopted; they also suggested
improvements to the arguments, some of which we adopted. Every adopted
correction and improvement was verified by the authors before being
incorporated, and none of them changed the statements of the results. We
disclose this use under both assisting in the writing of proofs and providing
critical ingredients for proving mathematical claims.

We have reviewed all AI-assisted work. All AI assistance was carried out under
the direction of the authors, and the resulting material was subsequently
checked and verified by two authors: every proof, including every AI-suggested
correction, was verified line by line independently of the tool's output;
AI-edited text was checked against the claims the authors intended to make. We
take responsibility for the final content of this work, including text, claims
or artifacts produced with the aid of generative AI.

\subsection*{Reproducibility statement}

The model architecture, Fourier cutoff, symbol degree, nonlinear-net inputs and the constraint parametrisation used in each experiment are listed in Appendix A.2.1, together with the data-generation procedure for the two synthetic problems and the splits for all four experiments. SEVIR and WildfireSpreadTS are publicly available; the preprocessing applied to each is described in Sections 3.3 and 3.4 and Appendix A.2.1. Code for the model, the baselines and the evaluation pipeline will be made publicly available.

\bibliographystyle{plainnat}
\bibliography{references}

\appendix
\section{Appendix}

\subsection{Proofs}

\subsubsection{Proofs of Section \ref{sec:stability}}
\label{app:stability}

Throughout this appendix, write
\[
  X:=\Op{V}_K,\qquad
  F:=F_K,\qquad
  \widetilde F:=\widetilde F_K,\qquad
  S_t:=\Op{S}^{K,\vartheta}_t,\qquad
  \widetilde S_t:=\Op{S}^{K,\widetilde\vartheta}_t.
\]
For vectors, $\|\cdot\|$ denotes the $L^2(\Omega)$ norm; for linear
operators, it denotes the induced operator norm.
Since $X$ is finite-dimensional, the locally Lipschitz maps
$F$ and $\widetilde F$ are Lipschitz on every bounded ball.
All Fourier multipliers considered below preserve real-valued
fields and commute with one another.

\begin{proof}[Proof of the estimate in
Theorem~\ref{thm:cross-step-stability}]
Set
\[
  B:=A-\widetilde A,\qquad
  h:=\tau_1-\tau_0,\qquad
  \varepsilon:=\varepsilon_{\rm obs}.
\]
Then $\|B\|\le 2M$.
We may choose
\begin{equation}
  \tau_\ast:=
  \min\left\{
    \Delta,\frac{\zeta}{2M},
    \frac{\log(R/r_0)}{3M}
  \right\},
  \qquad
  r_\Delta:=r_0\ee^{-M\Delta}.
  \label{eq:cross-step-radius-choice}
\end{equation}

We first control the difference between the linear operators.
For $t>0$ and $v\in X$, define
\begin{equation}
  Z_t(v):=
  \frac{
    S_t(\ee^{-t\widetilde A}v)
    -\widetilde S_t(\ee^{-t\widetilde A}v)
  }{t}.
  \label{eq:aligned-residual}
\end{equation}
Commutativity of $A$ and $\widetilde A$ gives the exact identity
\begin{equation}
  Z_t(v)
  =
  \frac{\ee^{tB}-I}{t}v
  +F(\ee^{tB}v)-\widetilde F(v).
  \label{eq:aligned-residual-identity}
\end{equation}
Consequently,
\begin{equation}
  Z_t(v)-2Z_t(v/2)+Z_t(0)
  =
  \mathcal N_F(\ee^{tB}v)
  -\mathcal N_{\widetilde F}(v),
  \label{eq:aligned-contrast}
\end{equation}
where
$\mathcal N_{\widetilde F}(v)
:=\widetilde F(v)-2\widetilde F(v/2)+\widetilde F(0)$.

Fix $w\in\Op{U}_{r_0}$ and set $v=\ee^{-\tau_0B}w$.
For $j=0,1$,
\[
  \|\ee^{-\tau_j\widetilde A}v\|
  \le
  \ee^{M\tau_j+2M\tau_0}r_0
  \le
  \ee^{3M\tau_\ast}r_0
  \le R.
\]
The same bound holds with $v$ replaced by $v/2$ or $0$.
The definition of $\varepsilon$ therefore implies
\[
  \|Z_{\tau_j}(v)\|,
  \ \|Z_{\tau_j}(v/2)\|,
  \ \|Z_{\tau_j}(0)\|
  \le\varepsilon,
  \qquad j=0,1.
\]
Subtracting Eq.~\eqref{eq:aligned-contrast} at the two observed
step sizes eliminates $\mathcal N_{\widetilde F}(v)$ and yields
\[
  \|\mathcal N_F(\ee^{hB}w)-\mathcal N_F(w)\|
  \le 8\varepsilon.
\]
Since $hB\in\mathcal M_K$ and
$\|hB\|\le2M\tau_\ast\le\zeta$,
Eq.~\eqref{eq:nonlinear-separation} gives
\begin{equation}
  \|B\|
  \le \frac{8\varepsilon}{\sigma h}.
  \label{eq:cross-step-generator-bound}
\end{equation}

To control the variation with the step size, define
\[
  Q_t(B):=\int_0^1 B\ee^{stB}\,ds,
  \qquad 0\le t\le\Delta.
\]
For $t>0$, this equals $(\ee^{tB}-I)/t$.
For $s,t\in[0,\Delta]$, the exponential integral identity gives
\begin{align}
  \|\ee^{tB}-\ee^{sB}\|
  &\le
  \|B\|\ee^{2M\Delta}|t-s|,
  \label{eq:cross-step-exponential-variation}\\
  \|Q_t(B)-Q_s(B)\|
  &\le
  \frac12\|B\|^2\ee^{2M\Delta}|t-s|
  \le
  M\|B\|\ee^{2M\Delta}|t-s|.
  \label{eq:cross-step-linear-variation}
\end{align}
For $v\in\Op{U}_{r_0}$ and $0\le t\le\Delta$,
\[
  \|\ee^{tB}v\|
  \le\ee^{2M\Delta}r_0
  \le\ee^{2M\Delta}R.
\]
Thus the Lipschitz bound $L_\Delta$ applies to the nonlinear
terms in Eq.~\eqref{eq:aligned-residual-identity}.
It follows that, for $s,t>0$,
\begin{equation}
  \|Z_t(v)-Z_s(v)\|
  \le D_\Delta\|B\||t-s|,
  \qquad
  D_\Delta:=r_0(M+L_\Delta)\ee^{2M\Delta}.
  \label{eq:cross-step-residual-variation}
\end{equation}
The term $\widetilde F(v)$ cancels in this difference.

Now fix $0<\delta\le\Delta$ and $u\in\Op{U}_{r_\Delta}$.
Set $v=\ee^{\delta\widetilde A}u$.
By Eq.~\eqref{eq:cross-step-radius-choice},
\[
  \|v\|\le\ee^{M\Delta}r_\Delta=r_0.
\]
Moreover,
\[
  \|\ee^{-\tau_j\widetilde A}v\|
  \le\ee^{M\tau_\ast}r_0\le R,
  \qquad j=0,1,
\]
so $\|Z_{\tau_j}(v)\|\le\varepsilon$.
Choose $\tau_j$ to be the nearer of the two observed step sizes.
Equations~\eqref{eq:cross-step-generator-bound}
and~\eqref{eq:cross-step-residual-variation} imply
\begin{align}
  \frac{\|S_\delta u-\widetilde S_\delta u\|}{\delta}
  &=\|Z_\delta(v)\| \notag\\
  &\le
  \varepsilon+
  D_\Delta\|B\||\delta-\tau_j| \notag\\
  &\le
  \varepsilon
  \left(
    1+\frac{8D_\Delta}{\sigma}
    \frac{|\delta-\tau_j|}{h}
  \right).
  \label{eq:cross-step-nearest-observation}
\end{align}
Write
\[
  d(\delta):=
  \min_{\tau\in[\tau_0,\tau_1]}|\delta-\tau|.
\]
The distance to the nearer endpoint satisfies
\[
  |\delta-\tau_j|\le \frac h2+d(\delta).
\]
Therefore,
\[
  1+\frac{8D_\Delta}{\sigma}
       \frac{|\delta-\tau_j|}{h}
  \le
  \left(1+\frac{4D_\Delta}{\sigma}\right)
  \left(1+\frac{2d(\delta)}{h}\right).
\]
Taking the supremum over $u\in\Op{U}_{r_\Delta}$ proves
Eq.~\eqref{eq:cross-step-stability}, with
\[
  C_\Delta:=1+\frac{4D_\Delta}{\sigma}.
\]
All constants have the dependence stated in the theorem.
\end{proof}

We next verify the nonlinear separation condition.
The following lemma includes the scalar condition without
derivative features.

\begin{lemma}[Derivative-free nonlinearities]
\label{lem:scalar-nonlinear-separation}
Suppose $F(v)=P_K[f(v)]$, where $f:\R\to\R$ is locally Lipschitz.
Let
\[
  \mathcal C_f(s):=f(s)-2f(s/2)+f(0).
\]
Assume that, for some $\nu>0$ and
$I=[s_-,s_+]$ with
$0<s_-<s_+<R/|\Omega|^{1/2}$,
\[
  |\mathcal C_f(s)-\mathcal C_f(t)|
  \ge\nu|s-t|,
  \qquad s,t\in I.
\]
Then Eq.~\eqref{eq:nonlinear-separation} holds for some
$0<r_0<R$ and $\sigma,\zeta>0$.
These constants can be chosen depending only on
$\Omega,K,R,I,\nu$.
\end{lemma}

\begin{proof}
Since $\mathcal C_f$ is continuous and injective on $I$,
it is strictly monotone there.
Hence there exists $\epsilon_f\in\{-1,1\}$ such that
\begin{equation}
  \epsilon_f
  \bigl(\mathcal C_f(s)-\mathcal C_f(t)\bigr)(s-t)
  \ge\nu|s-t|^2,
  \qquad s,t\in I.
  \label{eq:scalar-contrast-monotonicity}
\end{equation}

Let
\[
  c:=\frac{s_-+s_+}{2},
  \qquad
  d_I:=\frac{s_+-s_-}{2}.
\]
Finite dimensionality gives a constant $C_K>0$ such that
$\|v\|_{L^\infty}\le C_K\|v\|$ for $v\in X$.
Choose $\eta>0$ sufficiently small that
\[
  r_0:=c|\Omega|^{1/2}+\eta<R,
  \qquad C_K\eta\le d_I/4.
\]
Next choose $0<\zeta\le1/2$ such that
\[
  C_K(\ee^\zeta-1)r_0\le d_I/4.
\]

Fix $H\in\mathcal M_K$ with $\|H\|\le\zeta$.
For every real $q\in X$ with $\|q\|=1$, the fields
\[
  v_\pm:=c\mathbf 1\pm\eta q
\]
belong to $\Op{U}_{r_0}$ and take values in $I$.
Also,
\[
  \|\ee^Hv_\pm-v_\pm\|_{L^\infty}
  \le C_K(\ee^\zeta-1)r_0\le d_I/4,
\]
so $\ee^Hv_\pm$ take values in $I$.

For either of these fields, put $w=(\ee^H-I)v$.
Because $w\in X$ and
$\mathcal N_F(v)=P_K[\mathcal C_f(v)]$,
orthogonality of $P_K$ and
Eq.~\eqref{eq:scalar-contrast-monotonicity} imply
\begin{align*}
  \epsilon_f
  \langle\mathcal N_F(\ee^Hv)-\mathcal N_F(v),w\rangle
  &=
  \epsilon_f
  \int_\Omega
  \bigl(\mathcal C_f(\ee^Hv)-\mathcal C_f(v)\bigr)
  (\ee^Hv-v)\,dx\\
  &\ge\nu\|w\|^2.
\end{align*}
Consequently,
\[
  \|\mathcal N_F(\ee^Hv)-\mathcal N_F(v)\|
  \ge\nu\|(\ee^H-I)v\|.
\]

Let $T:=\ee^H-I$.
Choose a real unit vector $q\in X$ attaining $\|Tq\|=\|T\|$.
Since $Tv_+-Tv_-=2\eta Tq$,
\[
  \max\{\|Tv_+\|,\|Tv_-\|\}\ge\eta\|T\|.
\]
Finally, $\|H\|\le1/2$ gives
\[
  \|\ee^H-I-H\|
  \le\frac12\ee^{1/2}\|H\|^2
  \le\frac12\|H\|,
\]
and hence $\|T\|\ge\|H\|/2$.
Combining these estimates yields
\[
  \sup_{v\in\Op{U}_{r_0}}
  \|\mathcal N_F(\ee^Hv)-\mathcal N_F(v)\|
  \ge\frac{\nu\eta}{2}\|H\|.
\]
Thus Eq.~\eqref{eq:nonlinear-separation} holds with
$\sigma=\nu\eta/2$.
\end{proof}

The next lemma covers the quadratic transport nonlinearities
in the KdV and advection--diffusion examples.

\begin{lemma}[Quadratic transport nonlinearities]
\label{lem:transport-nonlinear-separation}
Let
$\Omega=\prod_{j=1}^{d_\Omega}(\R/\ell_j\Z)$,
with the Euclidean Fourier cutoff $|k|\le K$ and $K\ge2$.
Suppose
\[
  F(v)=-P_K[v(\mathbf a\cdot\nabla v)],
  \qquad \mathbf a\in\R^{d_\Omega}\setminus\{0\}.
\]
Then, for every $r_0>0$, there exist $\sigma,\zeta>0$
such that Eq.~\eqref{eq:nonlinear-separation} holds.
\end{lemma}

\begin{proof}
Write $\mathcal D:=\mathbf a\cdot\nabla$.
Since $F$ is homogeneous of degree two,
\[
  \mathcal N_F(v)
  =\frac12F(v)
  =-\frac14P_K\mathcal D(v^2).
\]
Its derivative satisfies
\begin{equation}
  D\mathcal N_F(v)[Hv]
  =-\frac12P_K\mathcal D(vHv).
  \label{eq:transport-contrast-derivative}
\end{equation}

We first show that
\[
  D\mathcal N_F(v)[Hv]=0\quad\text{for all real }v\in X
  \qquad\Longrightarrow\qquad H=0.
\]
Suppose the left-hand side holds.
Polarization of Eq.~\eqref{eq:transport-contrast-derivative}
gives
\[
  P_K\mathcal D(uHv+vHu)=0,
  \qquad u,v\in X.
\]
This bilinear identity extends to the complexification of $X$.
Let $b(k)$ denote the Fourier symbol of $H$, and set
$\alpha_j:=2\pi a_j/\ell_j$.
Testing the identity on the Fourier modes indexed by $p$ and $q$
gives
\begin{equation}
  \bigl(\boldsymbol\alpha\cdot(p+q)\bigr)
  \bigl(b(p)+b(q)\bigr)=0,
  \qquad
  |p|,|q|,|p+q|\le K.
  \label{eq:transport-mode-identity}
\end{equation}

Choose $i$ with $\alpha_i\ne0$.
Taking $p=q=e_i$ is allowed because $K\ge2$, and gives
$b(e_i)=0$.
Taking $p=0$, $q=e_i$ then gives $b(0)=0$.
For every retained mode $k$ with
$\boldsymbol\alpha\cdot k\ne0$, taking $p=0$, $q=k$ gives
\[
  b(k)=0.
\]

It remains to consider a nonzero retained mode $k$ satisfying
$\boldsymbol\alpha\cdot k=0$.
If $\alpha_i k_i\ne0$ for some $i$, choose
$p=-\operatorname{sgn}(k_i)e_i$.
Then
\[
  \boldsymbol\alpha\cdot p\ne0,
  \qquad
  |k+p|^2=|k|^2-2|k_i|+1\le |k|^2.
\]
Thus $p$ and $k+p$ are retained and have nonzero
$\boldsymbol\alpha$-frequency.
Equation~\eqref{eq:transport-mode-identity} with $q=k$
therefore implies $b(k)=0$.

Otherwise, $\alpha_i k_i=0$ for every $i$.
Choose $i,j$ with $\alpha_i\ne0$ and $k_j\ne0$.
Then $k_i=0$, $\alpha_j=0$, and $i\ne j$.
Set
$p=e_i-\operatorname{sgn}(k_j)e_j$.
Now
\[
  |p|=\sqrt2\le K,\qquad
  \boldsymbol\alpha\cdot p=\alpha_i\ne0,\qquad
  |k+p|^2=|k|^2-2|k_j|+2\le |k|^2.
\]
The same use of Eq.~\eqref{eq:transport-mode-identity}
gives $b(k)=0$.
Hence $b(k)=0$ for every retained mode, proving $H=0$.

By finite dimensionality and compactness, this injectivity implies
\[
  \kappa_K:=
  \min_{\substack{H\in\mathcal M_K\\\|H\|=1}}
  \ \sup_{\|v\|\le1}
  \|D\mathcal N_F(v)[Hv]\|>0.
\]
Homogeneity in $v$ and $H$ therefore gives
\begin{equation}
  \sup_{\|v\|\le r_0}
  \|D\mathcal N_F(v)[Hv]\|
  \ge\kappa_K r_0^2\|H\|.
  \label{eq:transport-infinitesimal-separation}
\end{equation}

To pass to finite perturbations, introduce the continuous symmetric
bilinear map
\[
  \mathcal B(v,w):=-\frac14P_K\mathcal D(vw),
  \qquad
  \mathcal N_F(v)=\mathcal B(v,v).
\]
Choose $C_{\mathcal B}>0$ such that
$\|\mathcal B(v,w)\|\le C_{\mathcal B}\|v\|\|w\|$.
For $\|H\|\le1$, set $T=\ee^H-I$.
The bounds
\[
  \|T\|\le\ee\|H\|,
  \qquad
  \|T-H\|\le\frac{\ee}{2}\|H\|^2
\]
and the quadratic identity give
\begin{align*}
  &\mathcal N_F(\ee^Hv)-\mathcal N_F(v)
    -D\mathcal N_F(v)[Hv]\\
  &\hspace{2em}
  =2\mathcal B(v,(T-H)v)+\mathcal B(Tv,Tv).
\end{align*}
Consequently,
\[
  \|\mathcal N_F(\ee^Hv)-\mathcal N_F(v)
    -D\mathcal N_F(v)[Hv]\|
  \le C_0\|v\|^2\|H\|^2,
  \qquad
  C_0:=C_{\mathcal B}(\ee+\ee^2).
\]
Together with Eq.~\eqref{eq:transport-infinitesimal-separation},
this yields
\[
  \sup_{\|v\|\le r_0}
  \|\mathcal N_F(\ee^Hv)-\mathcal N_F(v)\|
  \ge r_0^2(\kappa_K-C_0\|H\|)\|H\|.
\]
Taking
\[
  \zeta:=\min\{1,\kappa_K/(2C_0)\},
  \qquad
  \sigma:=\kappa_Kr_0^2/2
\]
proves the claim.
\end{proof}

For the KdV equation
\[
  u_t+a uu_x+b u_{xxx}=0,\qquad a\ne0,
\]
take
\[
  A=-b\partial_x^3,\qquad
  F(v)=-aP_K[vv_x].
\]
For the two-dimensional equation
\[
  u_t=\nu(u_{xx}+u_{yy})-u(u_x+u_y),
  \qquad \nu=0.05,
\]
on the periodic domain $[0,1]^2$, take
\[
  A=\nu(\partial_x^2+\partial_y^2),
  \qquad
  F(v)=-P_K[v(v_x+v_y)].
\]
In both cases, Lemma~\ref{lem:transport-nonlinear-separation}
verifies Eq.~\eqref{eq:nonlinear-separation} for $K\ge2$.
The linear operators are bounded on $X$, and the projected
quadratic nonlinearities are locally Lipschitz.
Thus both equations provide reference models satisfying the
hypotheses of Theorem~\ref{thm:cross-step-stability}.
The constants may depend on the fixed spatial resolution $K$.

Finally, we establish the sharpness assertion in
Theorem~\ref{thm:cross-step-stability}.
We show that the linear growth in
\[
  \frac{\min_{\tau\in[\tau_0,\tau_1]}|\delta-\tau|}
       {\tau_1-\tau_0}
\]
cannot, in general, be replaced by sublinear growth with constants
having the dependence stated in the theorem.

Take the reference model
\[
  A=0,\qquad F(v)=Q(v):=P_K[v^2].
\]
For $f(s)=s^2$, the scalar contrast is
$\mathcal C_f(s)=s^2/2$.
On every positive interval $I=[s_-,s_+]$,
\[
  |\mathcal C_f(s)-\mathcal C_f(t)|
  =\frac{s+t}{2}|s-t|
  \ge s_-|s-t|.
\]
Lemma~\ref{lem:scalar-nonlinear-separation} therefore verifies
the nonlinear separation condition.

Fix the reference model, the parameters in the theorem, and the
resulting constants $\tau_\ast$ and $r_\Delta$.
Choose
\[
  0<a<\tau_\ast/3,\qquad 0<|\gamma|\le M,
\]
and define a fixed comparison model by
\[
  \widetilde A=\gamma I,\qquad
  \widetilde F(v)=\ee^{-2\gamma a}Q(v)+\chi v,
  \qquad
  \chi:=\frac{\ee^{-\gamma a}-1}{a}.
\]
This is a model of the same form, with a locally Lipschitz
scalar nonlinearity.
A direct calculation gives
\begin{equation}
  D_t(v):=\frac{\widetilde S_t v-S_t v}{t}
  =\alpha(t)v+\beta(t)Q(v),
  \label{eq:cross-step-sharpness-residual}
\end{equation}
where
\[
  \alpha(t):=\frac{(1+t\chi)\ee^{\gamma t}-1}{t},
  \qquad
  \beta(t):=\ee^{2\gamma(t-a)}-1.
\]
Both coefficients vanish at $t=a$, so the two models agree
exactly at that step size.

Now take observed steps
\[
  \tau_0=a,\qquad \tau_1=a+h,
  \qquad 0<h<a/2,
\]
and let $h\downarrow0$, keeping both models fixed.
The functions $\alpha$ and $\beta$ are continuously
differentiable on $[a,3a/2]$, and $Q$ is bounded on
$\Op{U}_R$.
Consequently, for a constant $C_1$ independent of $h$,
\begin{equation}
  \varepsilon_{\rm obs}
  =
  \sup_{v\in\Op{U}_R}\|D_{a+h}(v)\|
  \le C_1h.
  \label{eq:cross-step-sharpness-observed}
\end{equation}

Consider the fixed inference step $\delta=2a\le\Delta$.
Since $Q(-v)=Q(v)$,
\[
  \frac{D_\delta(v)+D_\delta(-v)}{2}
  =\beta(\delta)Q(v).
\]
Testing the constant fields
$v=\pm r_\Delta/|\Omega|^{1/2}$ yields
\begin{equation}
  \sup_{v\in\Op{U}_{r_\Delta}}\|D_\delta(v)\|
  \ge
  |\ee^{2\gamma a}-1|
  \frac{r_\Delta^2}{|\Omega|^{1/2}}
  =:c_1>0.
  \label{eq:cross-step-sharpness-target}
\end{equation}
On the other hand,
\[
  q_h:=
  \frac{\min_{\tau\in[a,a+h]}|2a-\tau|}{h}
  =\frac{a-h}{h}\longrightarrow\infty.
\]
If the amplification factor in the theorem could be replaced
by $1+\psi(q_h)$ with $\psi(q)=o(q)$, then
Eq.~\eqref{eq:cross-step-sharpness-observed} would make its
right-hand side tend to zero:
\[
  C\varepsilon_{\rm obs}(1+\psi(q_h))
  \le CC_1h(1+\psi(q_h))
  \longrightarrow0.
\]
This contradicts Eq.~\eqref{eq:cross-step-sharpness-target}.
All model parameters and permitted constant dependencies remain
fixed as $h\downarrow0$.

The same argument with the fixed inference step $\delta=a/2$
establishes the corresponding sharpness for extrapolation below
the observed interval.
Thus the linear growth order is optimal in general; no optimality
of the numerical coefficient $2$ is asserted.

\subsubsection{Proofs of Section \ref{sec:identifiability}}
\label{app:identifiability}

Write $n:=d_\Omega$ and
$J_x\phi:=\Phi_\st[\phi](x)$.
Throughout this section, $\Op{S}_t$ denotes the original
one-step map
\begin{equation}
  \Op{S}_t[\phi;g,f](x)
  =
  (\ee^{tA}\phi)(x)
  +t f\bigl(J_x\ee^{tA}\phi,\eta(x)\bigr).
  \label{eq:ident-one-step-map}
\end{equation}
In particular, there is no projection of the final nonlinear
output in this definition.
We write the Fourier modes as $\ee^{\ii k\cdot x}$ on the
normalised torus.
On a periodic box with side lengths $\ell_j$, replace $k_j$
by $2\pi k_j/\ell_j$ in the differential symbols.
This invertible rescaling preserves the polynomial degrees and
the arguments below; the bandwidth bounds refer to integer
mode indices.

\begin{proof}[Proof of Lemma~\ref{lem:single-step}]
Let $v=\ee^{\tau_0A}\phi$.
Since $g_\gamma=g+\gamma$, the corresponding linear operator is
$A_\gamma=A+\gamma I$, and hence
\[
  \ee^{\tau_0A_\gamma}\phi=\varrho_\gamma v,
  \qquad
  J_x(\varrho_\gamma v)=\varrho_\gamma J_xv.
\]
Substitution into Eq.~\eqref{eq:ident-one-step-map} gives
\begin{align*}
  \Op{S}_{\tau_0}[\phi;g_\gamma,f_\gamma](x)
  &=
  \varrho_\gamma v(x)
  +\tau_0 f(J_xv,\eta(x))
  +\tau_0\chi_\gamma\varrho_\gamma v(x)\\
  &=
  v(x)+\tau_0 f(J_xv,\eta(x)),
\end{align*}
where the last equality uses
$\varrho_\gamma(1+\tau_0\chi_\gamma)=1$.
This proves Eq.~\eqref{eq:degeneracy}.
Repeated composition also gives equality of all rollouts
for which both iterates are defined.
\end{proof}

For the identifiability theorem, the following explicit choice
of bandwidth suffices:
\begin{equation}
  \begin{aligned}
    D&:=\max\{1,d\},\\
    N_\ast&:=
    \max\left\{
      (2\rho+1)D+2\rho+1,\,
      \rho+(\rho+2)(D+2\rho+1)
    \right\},\\
    T_\ast&:=\left\lceil\frac{N_\ast-1}{2}\right\rceil,\\
    K_0&:=
    \left\lceil
      \sqrt n\,\max\{\rho,DT_\ast\}
    \right\rceil.
  \end{aligned}
  \label{eq:ident-bandwidth}
\end{equation}
We first establish the two properties supplied by this threshold.

\begin{lemma}[Uniform realisation of local features]
\label{lem:ident-uniform-jets}
If $K\ge\rho\sqrt n$, then
$J_x:\Op{V}_K\to\R^{m_\st}$ is surjective for every $x\in\Omega$.
Moreover, there are linear right inverses
$\mathcal R_x:\R^{m_\st}\to\Op{V}_K$ satisfying
\[
  J_x\mathcal R_x=I,
  \qquad
  C_J:=\sup_{x\in\Omega}\|\mathcal R_x\|<\infty,
\]
where the operator norms use the Euclidean norm on the feature
space and the $L^2$ norm on $\Op{V}_K$.
\end{lemma}

\begin{proof}
Suppose a real linear functional annihilates the range of $J_x$.
Writing its coefficients as $a_\alpha$, this means
\[
  \sum_{|\alpha|\le\rho}
  a_\alpha\,\partial^\alpha\phi(x)=0,
  \qquad \phi\in\Op{V}_K.
\]
Complexifying this real linear identity and testing Fourier modes
gives
\[
  \sum_{|\alpha|\le\rho}a_\alpha(\ii k)^\alpha=0,
  \qquad k\in\Z^n,\quad |k|\le K.
\]
The retained frequencies contain the tensor grid
$\{0,\ldots,\rho\}^n$.
A polynomial of degree at most $\rho$ in each coordinate
that vanishes on this grid is identically zero, by repeated
univariate interpolation.
Thus all $a_\alpha$ vanish, proving surjectivity.

Choose a linear right inverse at one point.
Translating its output produces right inverses at all other
points, because differentiation commutes with translations.
Translations preserve the $L^2$ norm, so these right inverses
have the same operator norm.
\end{proof}

The next lemma supplies the finite-frequency rigidity needed
to identify the differential symbol.

\begin{lemma}[Finite-frequency rigidity]
\label{lem:ident-frequency-rigidity}
Let $H,P,Q$ be complex polynomials in $n$ real variables, with
degrees at most $d,\rho,\rho$, respectively, and suppose
$P(0)\ne0$.
Assume that, for every $k\in\Z^n$ with $|k|\le K$,
\[
  \ee^{H(k)}P(k)=Q(k),
  \qquad
  |\operatorname{Im}H(k)|<\pi/2.
\]
If $K\ge K_0$ with $K_0$ given by
Eq.~\eqref{eq:ident-bandwidth}, then $H$ is constant.
\end{lemma}

\begin{proof}
We first prove a univariate statement.
Let $h,p,q$ be complex polynomials of degrees at most
$D,\rho,\rho$, with $p$ and $q$ nonzero.
Suppose that at least $N_\ast$ distinct real points $t_j$ satisfy
\begin{equation}
  \ee^{h(t_j)}p(t_j)=q(t_j),
  \qquad
  |\operatorname{Im}h(t_j)|<\pi/2.
  \label{eq:ident-univariate-samples}
\end{equation}
We show that $h$ is constant.

Write $h(t)=a(t)+\ii b(t)$ for real $t$, where $a$ and $b$
are real polynomials.
Taking squared moduli in
Eq.~\eqref{eq:ident-univariate-samples} gives zeros of
\[
  R(t):=\ee^{2a(t)}|p(t)|^2-|q(t)|^2.
\]
Here $|p(t)|^2$ and $|q(t)|^2$ are real polynomials in $t$
of degrees at most $m:=2\rho$.

Suppose first that $a$ is nonconstant.
Differentiating $R$ exactly $m+1$ times eliminates the second
polynomial and gives
\[
  R^{(m+1)}(t)=\ee^{2a(t)}W(t),
  \qquad
  \deg W\le m+(m+1)(D-1).
\]
The polynomial $W$ is nonzero: at each differentiation,
the polynomial factor is transformed by
$U\mapsto U'+2a'U$, which maps every nonzero polynomial to a
nonzero polynomial when $a'$ is a nonzero polynomial.
Repeated Rolle's theorem therefore bounds the number of
distinct real zeros of $R$ by
\[
  m+(m+1)(D-1)+(m+1)
  =m+(m+1)D.
\]
This contradicts the definition of $N_\ast$.
Hence $a(t)=a_0$ is constant.
The polynomial
$|q(t)|^2-\ee^{2a_0}|p(t)|^2$ then has more zeros than its degree,
so
\begin{equation}
  |q(t)|^2=\ee^{2a_0}|p(t)|^2
  \qquad\text{for all real }t.
  \label{eq:ident-modulus-identity}
\end{equation}

Cancel the common polynomial factors in $q/p$ and write the
result as $U/V$, with $U$ and $V$ coprime.
For a polynomial $W$, let
$W^\#(z):=\overline{W(\overline z)}$.
Equation~\eqref{eq:ident-modulus-identity} implies the polynomial
identity
\[
  UU^\#=\ee^{2a_0}VV^\#.
\]
In particular, $\deg U=\deg V$.
Coprimality implies that $U$ divides $V^\#$, and therefore
\[
  U=C V^\#,\qquad |C|=\ee^{a_0}.
\]
The polynomial $V$ has no real roots, since any such root would
also be a root of $U$.
Consequently, the reduced rational function admits a smooth
argument on the real line:
\[
  \ee^{-a_0}\frac{U(t)}{V(t)}=\ee^{\ii\theta(t)}.
\]
If the roots of $V$ are $z_1,\ldots,z_\ell$, counted with
multiplicity, then $\ell\le\rho$, $\operatorname{Im}z_j\ne0$,
and
\begin{equation}
  \theta'(t)
  =
  -2\sum_{j=1}^{\ell}
  \frac{\operatorname{Im}z_j}
       {(t-\operatorname{Re}z_j)^2+(\operatorname{Im}z_j)^2}.
  \label{eq:ident-rational-phase}
\end{equation}
In particular,
\[
  \int_{\R}|\theta'(t)|\,dt\le2\pi\rho.
\]

At most $\rho$ of the sample points are zeros of $p$.
At each remaining point, Eq.~\eqref{eq:ident-univariate-samples}
gives
\[
  \theta(t_j)-b(t_j)\in2\pi\Z.
\]
The range of $\theta$ has width at most $2\pi\rho$, while
$b(t_j)\in(-\pi/2,\pi/2)$.
Thus at most $\rho+2$ distinct integers can occur in the
preceding identity.
By the definition of $N_\ast$, one integer $s$ occurs at least
$D+2\rho+1$ times.
The function
\[
  \theta(t)-b(t)-2\pi s
\]
therefore has at least that many distinct zeros.

By Eq.~\eqref{eq:ident-rational-phase},
$\theta'-b'$ is a rational function with a strictly positive
real denominator of degree at most $2\rho$ and a numerator
of degree at most $D+2\rho-1$.
Rolle's theorem supplies at least $D+2\rho$ distinct zeros
of this numerator, so it vanishes identically.
Hence $b'=\theta'$ on $\R$.
Since $\theta'(t)\to0$ as $|t|\to\infty$, the polynomial $b'$
must be zero.
Both $a$ and $b$ are constant, proving the univariate statement.

We now return to the multivariate polynomials.
The identity at $k=0$ implies $Q(0)\ne0$.
For each nonzero direction
$v\in\{0,\ldots,D\}^n$, restrict the polynomials to
\[
  h_v(t):=H(tv),\qquad
  p_v(t):=P(tv),\qquad
  q_v(t):=Q(tv).
\]
The polynomials $p_v$ and $q_v$ are nonzero because their
values at zero are nonzero.
For every integer $t$ with $|t|\le T_\ast$,
\[
  |tv|\le T_\ast D\sqrt n\le K.
\]
Thus these restrictions satisfy the univariate hypotheses
at $2T_\ast+1\ge N_\ast$ distinct points.
It follows that $H(tv)$ is constant in $t$, and in particular
$H(v)=H(0)$.

Therefore $H-H(0)$ vanishes on the tensor grid
$\{0,\ldots,D\}^n$.
Its degree in each coordinate is at most $D$, so repeated
univariate interpolation gives $H\equiv H(0)$.
\end{proof}

\begin{proof}[Proof of Theorem~\ref{thm:identifiability}]
We first prove part~{\rm(i)}.
Choose $x_0$ as in its nondegeneracy assumption and set
$\eta_0:=\eta(x_0)$.
Differentiate the equality of the prediction maps at the zero
input and evaluate at $x_0$.
For every real $\psi\in\Op{V}_K$, this gives
\begin{align*}
  &(\ee^{\tau_0A}\psi)(x_0)
  +\tau_0
  \sum_{|\alpha|\le\rho}
  \partial_{z_\alpha}f(0,\eta_0)
  \,\partial^\alpha(\ee^{\tau_0A}\psi)(x_0)\\
  &\qquad=
  (\ee^{\tau_0A'}\psi)(x_0)
  +\tau_0
  \sum_{|\alpha|\le\rho}
  \partial_{z_\alpha}f'(0,\eta_0)
  \,\partial^\alpha(\ee^{\tau_0A'}\psi)(x_0).
\end{align*}
Define the polynomials
\[
  P(k):=
  1+\tau_0\sum_{|\alpha|\le\rho}
  \partial_{z_\alpha}f(0,\eta_0)(\ii k)^\alpha,
  \qquad
  Q(k):=
  1+\tau_0\sum_{|\alpha|\le\rho}
  \partial_{z_\alpha}f'(0,\eta_0)(\ii k)^\alpha.
\]
Complexifying the linear identity and testing Fourier modes
yields
\begin{equation}
  \ee^{\tau_0g(k)}P(k)
  =
  \ee^{\tau_0g'(k)}Q(k),
  \qquad |k|\le K.
  \label{eq:ident-linearised-symbols}
\end{equation}
The nondegeneracy assumption states precisely that
\[
  P(0)=1+\tau_0\partial_{z_0}f(0,\eta_0)\ne0.
\]

Apply Lemma~\ref{lem:ident-frequency-rigidity} to
\[
  H(k):=\tau_0\bigl(g(k)-g'(k)\bigr).
\]
Indeed, Eq.~\eqref{eq:ident-linearised-symbols} gives
$\ee^{H(k)}P(k)=Q(k)$, and the two anti-aliasing bounds imply
\[
  |\operatorname{Im}H(k)|
  \le
  \tau_0|\operatorname{Im}g(k)|
  +\tau_0|\operatorname{Im}g'(k)|
  <\pi/2.
\]
Hence $g-g'$ is constant.
Since both differential symbols have real coefficients,
their values at zero are real.
Thus, for some $\gamma\in\R$,
\[
  g'=g+\gamma,\qquad A'=A+\gamma I.
\]
Equality of the positive-order polynomial coefficients follows
immediately.

It remains to identify the nonlinear function on a common
feature neighbourhood.
Let $\mathcal R_x$ and $C_J$ be as in
Lemma~\ref{lem:ident-uniform-jets}.
For any step size $t$ at which the prediction maps agree
on the stated input ball, define
\[
  r_t:=
  \frac{r}{2C_J\|\ee^{-tA'}\|_{\Op{V}_K\to\Op{V}_K}}.
\]
This is positive because the exponential is invertible on the
finite-dimensional space $\Op{V}_K$.
For $|z_\st|<r_t$, take
\[
  \phi:=\ee^{-tA'}\mathcal R_x z_\st.
\]
Then $\|\phi\|_{L^2}<r$, and
\[
  J_x\ee^{tA'}\phi=z_\st,
  \qquad
  J_x\ee^{tA}\phi=\ee^{-\gamma t}z_\st.
\]
Equality of the two prediction maps at $x$ therefore gives
\begin{equation}
  f'(z_\st,\eta(x))
  =
  f(\ee^{-\gamma t}z_\st,\eta(x))
  +\frac{\ee^{-\gamma t}-1}{t}\,z_0,
  \qquad |z_\st|<r_t.
  \label{eq:ident-readout-at-step}
\end{equation}
The radius $r_t$ is independent of $x$.
Since $E=\eta(\Omega)$, this identity holds for every
$z_\eta\in E$.
Taking $t=\tau_0$ and $r_\ast=r_{\tau_0}$ proves
$f'=f_\gamma$ on $B_{r_\ast}(0)\times E$, completing part~{\rm(i)}.

For part~{\rm(ii)}, put
\[
  a_j:=\ee^{-\gamma\tau_j},
  \qquad
  \chi_j:=\frac{a_j-1}{\tau_j},
  \qquad j=0,1.
\]
Equation~\eqref{eq:ident-readout-at-step} applies at both
observed step sizes.
Thus, on a common neighbourhood of the origin and for every
$z_\eta\in E$,
\begin{equation}
  f(a_0z_\st,z_\eta)+\chi_0z_0
  =
  f(a_1z_\st,z_\eta)+\chi_1z_0.
  \label{eq:ident-two-step-scaling}
\end{equation}

Suppose $\gamma\ne0$, and choose $z_\eta^\ast$ as in the
non-affinity assumption.
Since $\tau_0\ne\tau_1$,
\[
  \lambda:=\frac{a_0}{a_1}>0,\qquad \lambda\ne1.
\]
Setting $w=a_1z_\st$ in
Eq.~\eqref{eq:ident-two-step-scaling} gives
\[
  f(\lambda w,z_\eta^\ast)-f(w,z_\eta^\ast)
  =b\,w_0,
  \qquad
  b:=\frac{\chi_1-\chi_0}{a_1},
\]
for all $w$ sufficiently close to zero.
Define
\[
  H_\ast(w):=
  f(w,z_\eta^\ast)-\frac{b}{\lambda-1}w_0.
\]
Then $H_\ast(\lambda w)=H_\ast(w)$ locally.
Writing
$\mu:=\min\{\lambda,\lambda^{-1}\}\in(0,1)$,
this implies, on a sufficiently small ball,
\[
  H_\ast(\mu w)=H_\ast(w).
\]
Iteration and continuity at zero yield
\[
  H_\ast(w)=H_\ast(\mu^m w)\longrightarrow H_\ast(0)
  \qquad(m\to\infty).
\]
Consequently,
\[
  f(w,z_\eta^\ast)
  =
  f(0,z_\eta^\ast)+\frac{b}{\lambda-1}w_0
\]
on a neighbourhood of the origin.
This contradicts the assumed non-affinity.
Therefore $\gamma=0$.
Part~{\rm(i)} now gives $g'=g$ and $f'=f$ on its original
feature domain $B_{r_\ast}(0)\times E$.
\end{proof}

\subsection{Numerical Experiments}

\subsubsection{Main ENNO hyperparameters}
\label{app:hyper}

Tables~\ref{tab:arch} and~\ref{tab:data} list, for each of the four experiments, the
architecture of ENNO and the data it was trained and evaluated on. Throughout, $K$ is the
Euclidean cutoff on integer mode indices, $|k|\le K$, as in Section~2.3.1; $d$ is the degree
of the symbol $g$ in Eq.~(6); $\rho$ is the order of the jet entering the nonlinear map $f$ in
Eq.~(8); and $v$ denotes the linearly propagated field to which $f$ is applied in Eq.~\eqref{eq:onestep}.
The coefficients of the even monomials in $\operatorname{Re} g$ (Section~2.2) are parametrised
as $-\operatorname{softplus}(\theta)$, so that $\operatorname{Re} g\le c_0$ holds by
construction; the odd-order coefficients and $c_0$ are unconstrained. One consequence is that a
constrained coefficient whose true value is zero cannot reach zero exactly and settles at the
floor of the parametrisation; this is the case for $c_2$ in Table~2. The unconstrained coefficients are initialised from $\mathcal N(0,0.01^2)$.
The nonlinear map $f$ and, where they are not the identity, the encoder $E$ and the decoder
$R$ are multilayer perceptrons with Tanh activations; the output layer of $f$ is
zero-initialised, so that training starts from the purely linear flow.

ENNO is trained with Adam for at most 3{,}000 epochs (the cap applied to all neural models in
Section~3), and its checkpoint is
selected by early stopping on the validation set (patience 15 for the synthetic problems,
evaluated on the noised validation trajectories, and 200 for the real data; for
WildfireSpreadTS the criterion is validation AUC-PR, Section~3.4). All reported results come
from a single training seed per configuration (per fold for WildfireSpreadTS). Relative $L^2$ errors on the synthetic problems are computed against
the clean, noise-free test trajectories; the targets at $t_{1/2}$ come from the same solver
runs at the finer output interval (Table~\ref{tab:data}).

\begin{table}[h]
\centering
\small
\caption{Architecture of ENNO in the four experiments.}
\label{tab:arch}
\begin{tabularx}{\textwidth}{L{3.3cm} Y Y Y Y}
\toprule
 & KdV & Advection--diffusion & SEVIR (VIL) & WildfireSpreadTS \\
\midrule
Latent channels $M$
 & 1 & 1 & 20 & 4 ($T_{\mathrm{in}}=1$) / 20 ($T_{\mathrm{in}}=5$) \\
History length (frames)
 & 1 & 1 & 10 & $T_{\mathrm{in}}\in\{1,5\}$ \\
Encoder $E$ / decoder $R$
 & identity (learnable variants in Table~\ref{tab:kdv_ablation})
 & identity
 & pointwise MLPs $\mathbb{R}^{10}\!\to\!\mathbb{R}^{20}$ and $\mathbb{R}^{20}\!\to\!\mathbb{R}$, 5 hidden layers $\times$ 256
 & same design, $\mathbb{R}^{T_{\mathrm{in}}}\!\to\!\mathbb{R}^{M}$ and $\mathbb{R}^{M}\!\to\!\mathbb{R}$, 5 hidden layers $\times$ 256 \\
Cutoff $K$ ($|k|\le K$)\textsuperscript{a}
 & 64 & 64 & 64 & 32 \\
Symbol degree $d$
 & 3 (ablation over $1$--$4$, Table~1)
 & 2 (ablation over $1$--$3$, Table~4)
 & 6
 & 6 \\
Constant term $c_0$
 & trained & trained & fixed at 0 & fixed at 0 \\
Jet order $\rho$
 & 2 & 2 & 1 & 1 \\
Inputs of $f$
 & $v,\,v_x,\,v_{xx}$
  & $v,\,v_x,\,v_y$, $v_{xx},\,v_{yy},\,v_{xy}$
 & $v,\,v_x,\,v_y$
 & $v,\,v_x,\,v_y$ and 25 environmental channels $\eta$ \\
Size of $f$
 & 4 hidden layers $\times$ 128 & 4 $\times$ 128 & 4 $\times$ 256 & 4 $\times$ 256 \\
\bottomrule
\end{tabularx}
\vspace{2pt}
\begin{flushleft}\footnotesize
\textsuperscript{a}Euclidean cutoff on integer mode indices, matching $V_K$ in Section~2.3.1.
On the 128-point and $128\times128$ grids, $K=64$ is the Nyquist index (in two dimensions the
corners of the square spectrum beyond radius 64 are dropped); for WildfireSpreadTS, $K=32$ is a
genuine low-pass.
\end{flushleft}
\end{table}

\begin{table}[h]
\centering
\small
\caption{Data used in the four experiments.}
\label{tab:data}
\begin{tabularx}{\textwidth}{L{3.3cm} Y Y Y Y}
\toprule
 & KdV & Advection--diffusion & SEVIR & WildfireSpreadTS \\
\midrule
Domain and grid
 & $[0,10]$ periodic, 128 points
 & $[0,1]^2$ periodic, $128\times128$
 & $384\times384$\,km$^2$; VIL on $384^2$ pixels, average-pooled to $128^2$
 & daily fire maps at 375\,m, padded to $360\times320$ \\
Source
 & Fourier pseudospectral in space, DOP853 in time (rtol $10^{-8}$);
   10{,}001 frames over $T=1$, every 10th kept
 & explicit Euler, $\Delta t=10^{-5}$, central differences in space;
   5{,}001 frames over $T=0.5$, every 10th kept
 & real radar, 49 frames per event
 & real observations, 2018--2021 \\
Initial conditions
 & 3 sine modes (amplitude $\mathcal U(-1,1)/m^2$ for mode $m$) plus 1 or 2 solitons (amplitude $\mathcal U(1.5,3.5)$)
 & sums of 1--3 Gaussian bumps (amplitude $\mathcal U(0.5,1.5)$, $\sigma\sim\mathcal U(0.05,0.25)$) and low-mode Fourier fields
 & --- & --- \\
Observation interval
 & $10^{-3}$ & $10^{-3}$ & 5\,min & 1\,day \\
Training step sizes $\tau_1,\tau_2$
 & $10^{-3},\;2\times10^{-3}$ & $10^{-3},\;2\times10^{-3}$ & 5, 10\,min & 1, 2\,days \\
Targets at $t_{1/2}$
 & every 5th frame of the same runs & every 5th frame of the same runs & --- & --- \\
Train / val / test
 & 2{,}030 / 50 / 200 trajectories & 100 / 10 / 25 trajectories & 11{,}159 / 200 / 500 events & four leave-one-year-out folds \\
Noise (train and val, all methods)
 & i.i.d.\ Gaussian, 5\% of each trajectory's std, fixed seed; test trajectories clean
 & same & none & none \\
\bottomrule
\end{tabularx}
\end{table}

\subsubsection{Ablation results of the KdV example} \label{app:kdv}
The encoder and decoder of Eq.~\eqref{eq:latentrollout} are identity maps in
the configuration reported above, so the model acts directly on $u$.
Table~\ref{tab:kdv_ablation} makes them learnable, with $M$ latent fields, at each symbol degree. Where the degree is below the true
order, the learnable variant with $M\ge4$ is lower at every target time, and
the gain saturates between $M=4$ and $M=16$: at $d=1$ it reaches $2.449\%$
at $t_{10}$ against $4.189\%$ for the identity.  At $d\ge3$, where the true
symbol is already in the class, the pattern splits: the identity is lower at
$t_1$ and $t_5$, by a factor of $2.7$ at $t_1$, while the learnable variant
grows more slowly and ends slightly lower at $t_{10}$.  We report the identity
configuration, which is the more accurate of the two over most of the horizon.

\begin{table}[!ht]
  \centering
  \small
  \setlength{\tabcolsep}{8pt}
  \caption{KdV: encoder/decoder ablation. Relative $L^2$ error (\%) at three
  target frames. ``Identity'' is the parameter-free configuration reported in
  the main text; the other rows make them learnable, with $M$ latent fields. Best in each block in bold.}
  \label{tab:kdv_ablation}
  \begin{tabular}{@{}lrrr@{}}
    \toprule
    Variant & $t_{1}$ & $t_{5}$ & $t_{10}$ \\
    \midrule
    $d=1$, identity & 2.898 & 3.683 & 4.189 \\
    $d=1$, $M=1$    & 3.069 & 2.629 & 3.276 \\
    $d=1$, $M=4$    & \textbf{1.866} & 2.317 & 2.468 \\
    $d=1$, $M=16$   & 1.877 & \textbf{2.289} & \textbf{2.449} \\
    \addlinespace
    $d=2$, identity & 2.181 & 2.700 & 2.994 \\
    $d=2$, $M=1$    & 1.990 & 2.454 & 2.856 \\
    $d=2$, $M=4$    & \textbf{1.532} & \textbf{2.233} & \textbf{2.529} \\
    $d=2$, $M=16$   & 1.535 & 2.235 & \textbf{2.529} \\
    \addlinespace
    $d=3$, identity & \textbf{0.313} & \textbf{0.721} & 1.115 \\
    $d=3$, $M=1$    & 0.852 & {0.915} & \textbf{0.994} \\
    $d=3$, $M=4$    & 0.846 & 0.939 & 1.028 \\
    $d=3$, $M=16$   & 0.842 & 0.935 & 1.014 \\
    \addlinespace
    $d=4$, identity & \textbf{0.318} & \textbf{0.738} & 1.152 \\
    $d=4$, $M=1$    & 0.846 & {0.920} & \textbf{1.007} \\
    $d=4$, $M=4$    & 0.840 & 0.968 & 1.086 \\
    $d=4$, $M=16$   & 0.832 & 0.963 & 1.098 \\
    \bottomrule
  \end{tabular}
\end{table}
\label{app:sevir}
\end{document}

%% file: math_commands.tex
\usepackage{amsmath,amsfonts,bm}

\def\1{\bm{1}}

\DeclareMathAlphabet{\mathsfit}{\encodingdefault}{\sfdefault}{m}{sl}
\SetMathAlphabet{\mathsfit}{bold}{\encodingdefault}{\sfdefault}{bx}{n}

\newcommand{\R}{\mathbb{R}}



%% file: references.bib
@book{brunton2019data,
  title     = {Data-Driven Science and Engineering: Machine Learning, Dynamical Systems, and Control},
  author    = {Brunton, Steven L. and Kutz, J. Nathan},
  publisher = {Cambridge University Press},
  year      = {2019}
}

@article{brunton2016discovering,
  title   = {Discovering governing equations from data by sparse identification of nonlinear dynamical systems},
  author  = {Brunton, Steven L. and Proctor, Joshua L. and Kutz, J. Nathan},
  journal = {Proceedings of the National Academy of Sciences},
  volume  = {113},
  number  = {15},
  pages   = {3932--3937},
  year    = {2016}
}

@article{chen2025due,
  title   = {{DUE}: A deep learning framework and library for modeling unknown equations},
  author  = {Chen, Junfeng and Wu, Kailiang and Xiu, Dongbin},
  journal = {SIAM Review},
  volume  = {67},
  number  = {4},
  pages   = {873--902},
  year    = {2025}
}

@article{fasel2022esindy,
  title   = {Ensemble-{SINDy}: Robust sparse model discovery in the low-data, high-noise limit, with active learning and control},
  author  = {Fasel, Urban and Kutz, J. Nathan and Brunton, Bingni W. and Brunton, Steven L.},
  journal = {Proceedings of the Royal Society A},
  volume  = {478},
  number  = {2260},
  pages   = {20210904},
  year    = {2022}
}

@inproceedings{Gerard2023WildfireSpreadTSAD,
  title     = {{WildfireSpreadTS}: A dataset of multi-modal time series for wildfire spread prediction},
  author    = {Gerard, Sebastian and Zhao, Yu and Sullivan, Josephine},
  booktitle = {Advances in Neural Information Processing Systems},
  volume    = {36},
  year      = {2023}
}

@inproceedings{herde2024poseidon,
  title     = {Poseidon: Efficient foundation models for {PDEs}},
  author    = {Herde, Maximilian and Raoni{\'c}, Bogdan and Rohner, Tobias and K{\"a}ppeli, Roger and Molinaro, Roberto and de B{\'e}zenac, Emmanuel and Mishra, Siddhartha},
  booktitle = {Advances in Neural Information Processing Systems},
  volume    = {37},
  pages     = {72525--72624},
  year      = {2024}
}

@inproceedings{hou2026cfo,
  title     = {{CFO}: Learning continuous-time {PDE} dynamics via flow-matched neural operators},
  author    = {Hou, Xianglong and Huang, Xinquan and Perdikaris, Paris},
  booktitle = {International Conference on Learning Representations},
  year      = {2026}
}

@article{kovachki2023neural,
  title   = {Neural operator: Learning maps between function spaces with applications to {PDEs}},
  author  = {Kovachki, Nikola and Li, Zongyi and Liu, Burigede and Azizzadenesheli, Kamyar and Bhattacharya, Kaushik and Stuart, Andrew and Anandkumar, Anima},
  journal = {Journal of Machine Learning Research},
  volume  = {24},
  number  = {89},
  pages   = {1--97},
  year    = {2023}
}

@article{li2025nno,
  title   = {Nemytskii neural operator: {A} nonlinear model reduction method for parametrized partial differential equations},
  author  = {Li, Jingye and Bespalov, Alex and Li, Jinglai},
  journal = {arXiv preprint},
  note    = {arXiv:2511.07684},
  year    = {2025}
}

@inproceedings{li2021fourier,
  title     = {Fourier Neural Operator for Parametric Partial Differential Equations},
  author    = {Li, Zongyi and Kovachki, Nikola and Azizzadenesheli, Kamyar and Liu, Burigede and Bhattacharya, Kaushik and Stuart, Andrew and Anandkumar, Anima},
  booktitle = {International Conference on Learning Representations},
  year      = {2021}
}

@inproceedings{long2018pde,
  title     = {{PDE-Net}: Learning {PDEs} from Data},
  author    = {Long, Zichao and Lu, Yiping and Ma, Xianzhong and Dong, Bin},
  booktitle = {Proceedings of the 35th International Conference on Machine Learning},
  series    = {Proceedings of Machine Learning Research},
  volume    = {80},
  pages     = {3208--3216},
  year      = {2018}
}

@article{lu2021learning,
  title   = {Learning nonlinear operators via {DeepONet} based on the universal approximation theorem of operators},
  author  = {Lu, Lu and Jin, Pengzhan and Pang, Guofei and Zhang, Zhongqiang and Karniadakis, George Em},
  journal = {Nature Machine Intelligence},
  volume  = {3},
  number  = {3},
  pages   = {218--229},
  year    = {2021}
}

@article{lu2022comprehensive,
  title   = {A comprehensive and fair comparison of two neural operators (with practical extensions) based on {FAIR} data},
  author  = {Lu, Lu and Meng, Xuhui and Cai, Shengze and Mao, Zhiping and Goswami, Somdatta and Zhang, Zhongqiang and Karniadakis, George Em},
  journal = {Computer Methods in Applied Mechanics and Engineering},
  volume  = {393},
  pages   = {114778},
  year    = {2022}
}

@article{nayak2026ti,
  title   = {{TI-DeepONet}: Learnable time integration for stable long-term extrapolation},
  author  = {Nayak, Dibyajyoti and Goswami, Somdatta},
  journal = {Computer Methods in Applied Mechanics and Engineering},
  volume  = {456},
  pages   = {118960},
  year    = {2026}
}

@article{rudy2017data,
  title   = {Data-driven discovery of partial differential equations},
  author  = {Rudy, Samuel H. and Brunton, Steven L. and Proctor, Joshua L. and Kutz, J. Nathan},
  journal = {Science Advances},
  volume  = {3},
  number  = {4},
  pages   = {e1602614},
  year    = {2017}
}

@inproceedings{seidman2022nomad,
  title     = {{NOMAD}: Nonlinear manifold decoders for operator learning},
  author    = {Seidman, Jacob and Kissas, Georgios and Perdikaris, Paris and Pappas, George J.},
  booktitle = {Advances in Neural Information Processing Systems},
  volume    = {35},
  pages     = {5601--5613},
  year      = {2022}
}

@inproceedings{veillette2020sevir,
  title     = {{SEVIR}: A Storm Event Imagery Dataset for Deep Learning Applications in Radar and Satellite Meteorology},
  author    = {Veillette, Mark and Samsi, Siddharth and Mattioli, Chris},
  booktitle = {Advances in Neural Information Processing Systems},
  volume    = {33},
  year      = {2020}
}

@article{wang2021learning,
  title   = {Learning the solution operator of parametric partial differential equations with physics-informed {DeepONets}},
  author  = {Wang, Sifan and Wang, Hanwen and Perdikaris, Paris},
  journal = {Science Advances},
  volume  = {7},
  number  = {40},
  pages   = {eabi8605},
  year    = {2021}
}

@article{wooster2005,
  title   = {Retrieval of biomass combustion rates and totals from fire radiative power observations: {FRP} derivation and calibration relationships between biomass consumption and fire radiative energy release},
  author  = {Wooster, M. J. and Roberts, G. and Perry, G. L. W. and Kaufman, Y. J.},
  journal = {Journal of Geophysical Research: Atmospheres},
  volume  = {110},
  number  = {D24},
  pages   = {D24311},
  year    = {2005}
}

@article{xiong2024koopman,
  title   = {{Koopman} neural operator as a mesh-free solver of non-linear partial differential equations},
  author  = {Xiong, Wei and Huang, Xiaomeng and Zhang, Ziyang and Deng, Ruixuan and Sun, Pei and Tian, Yang},
  journal = {Journal of Computational Physics},
  volume  = {513},
  pages   = {113194},
  year    = {2024}
}
